\documentclass[12pt]{article}

\usepackage{graphicx}
\usepackage{amssymb}
 
\usepackage{JASA_manu}

\usepackage{natbib}
\usepackage{amsfonts}
\usepackage{amsmath}
\usepackage{rotating}
\usepackage{booktabs}
\usepackage{caption}
\usepackage{longtable}
\usepackage{ctable}
\usepackage{mathrsfs}
\usepackage{graphicx}
\usepackage{psfrag,epsfig,subfigure}
\usepackage{bm}
\usepackage{algorithm}
\usepackage{algorithmic}

\usepackage{color}
\newcommand{\bY}{\mathbf{Y}}
\newcommand{\bX}{\mathbf{X}}
\newcommand{\bB}{\mathbf{B}}
\newcommand{\bE}{\mathbf{E}}
\newcommand{\bg}{\mathbf{g}}
\newcommand{\bh}{\mathbf{h}}
\newcommand{\bl}{\mathbf{l}}
\newcommand{\by}{\mathbf{y}}
\newcommand{\bx}{\mathbf{x}}
\newcommand{\br}{\mathbf{r}}

\def\bSig\mathbf{\Sigma}

\begin{document}

\title{Structured Input-Output Lasso, \\
with Application to eQTL Mapping, and a Thresholding Algorithm for Fast Estimation}
\author{Seunghak Lee$^{*}$  \\
Eric P. Xing$^{**}$ \\ 
School of Computer Science \\ 
Carnegie Mellon University, Pittsburgh, PA, U.S.A. \\ 
$^{*}$email: \texttt{seunghak@cs.cmu.edu}\\
$^{**}$email: \texttt{epxing@cs.cmu.edu} }

\maketitle

\newpage
\begin{center}
\textbf{Abstract}
\end{center}
We consider the problem of learning a high-dimensional multi-task regression model, under sparsity constraints induced by presence of grouping structures on the input covariates and on the output predictors. This problem is primarily motivated by expression quantitative trait locus (eQTL) mapping, of which the goal is to discover genetic variations in the genome (inputs) that influence the expression levels of multiple co-expressed genes (outputs), either epistatically, or pleiotropically, or both. A structured input-output lasso (SIOL) model based on an intricate $\ell_1/\ell_2$-norm penalty over the regression coefficient matrix is employed to enable discovery of complex sparse input/output relationships; and a highly efficient new optimization algorithm called hierarchical group thresholding (HiGT) is developed to solve the resultant non-differentiable, non-separable, and ultra high-dimensional optimization problem. We show on both simulation and on a yeast eQTL dataset that our model leads to significantly better recovery of the structured sparse relationships between the inputs and the outputs, and our algorithm significantly outperforms other optimization techniques under the same model. Additionally, we propose a novel approach for efficiently and effectively detecting input interactions by exploiting the prior knowledge available from biological experiments.

\vspace*{.3in}

\noindent\textsc{Keywords}: {Group Lasso, Multi-task Lasso, 
Epistasis, Pleiotrophy, Genome-wide association studies, eQTL mapping, Genetic interaction network} 

\newpage

\section{Introduction}
\label{sec:intro}
In this paper, we consider the problem of learning a functional mapping
from a high-dimensional input space with structured sparsity to a multivariate output space where responses are coupled (therefore making the estimator doubly structured), with an application for detecting genomic loci
affecting gene expression levels, a problem known as expression quantitative trait loci (eQTL) mapping.
In particular, we are interested in exploiting the structural information 
on both the input and output space jointly to improve the accuracy for identifying 
a small number of input variables relevant to the outputs
among a large number of candidates. 
When the input or output variables are highly correlated among themselves,
multiple related inputs may synergistically influence 
the same outputs,
and multiple related outputs may be synergistically influenced by the same inputs.

The primary motivation for our work comes from the problem of genome-wide association (GWA) mapping of eQTLs in computational genomics \cite{kendziorski2006statistical}, of which 
the goal 
is to detect the genetic variations, often single nucleotide polymorphisms (SNPs), across the whole genome 
that perturb the expression levels of genes, given the data of SNP genotypes 
and microarray gene-expression traits of a study cohort. 
One of the main challenges of this problem 
is that, typically the sample size is very small (e.g. $\sim 1000$), whereas there are a very large number of SNPs (e.g. $\sim$ 500,000) and expression traits (e.g., $\sim$ 10,000).
Furthermore, there have been numerous evidences that multiple genetic variations may interact with
each other (a.k.a., epistasis) \cite{segre2005modular,cho1998identification,badano2005dissection}, and the same genetic variation(s) can influence multiple genes 
(a.k.a., pleiotropy) \cite{stearns2010one,ducrest2008pleiotropy}. 
However, prior knowledge of such information 
implying complex association 
structures are difficult to exploit in standard 
statistical analysis of GWA mapping \cite{wang2010analysing,moore2010bioinformatics}.
To enhance the statistical power for mapping of eQTLs, 
it is desirable to incorporate biological knowledge of genome and transcriptome structures into the model to guide the search for true eQTLs.
In this article, we focus on developing a model 
which can make use of structural information on both input (SNPs) and output (gene expressions) sides. In particular, we consider biological knowledge about group associations as structural information. If there exist group behaviors among the covariates in the high-dimensional input $\bX$, for example, multiple genetically coupled SNPs 
(e.g., in linkage disequilibrium) can jointly affect a single trait \cite{wang2010analysing}, 
such group information is called an input structure;
if multiple variables in the high-dimensional output $\bY$ are jointly under the influence of a similar set of input covariates, for example, a single SNP can affect multiple functionally coupled traits (e.g., genes in the same pathway or operon) \cite{kim-xing-PLoSG09}, 
such group information is called an output structure.
The problem of GWA mapping of eQTLs can be formulated as a model selection problem under a multitask regression model $\bY=\bB \bX$ with structured sparsity, 
where the resultant non-zero elements in the regression coefficient matrix $\bB$ expose the identities of the eQTLs and their associated traits.

Variants of this problem have been widely studied in the recent high-dimensional inference and variable selection literature, and various penalty-based or Bayesian approaches for learning a shared sparsity pattern among either multiple inputs or multiple outputs in a regression model have been proposed 
\cite{tibshirani2005sparsity,negahban2011simultaneous,han2010multi,li2010bayesian}. 
Depending on the type of structural constraints, different penalty functions
have been previously considered, including
mixed-norm (group-lasso) penalty for a simple grouping structure 
\cite{yuan2006model,zhao2009composite}, 
tree-guided group-lasso 
penalty for a tree structure \cite{kim2009tree}, 
or graph-guided fused lasso for a graph structure \cite{kim-xing-PLoSG09}.
Most previous approaches, however,
considered either only the input structural constraints 
or only the output structural constraints, but not both. 
There have been a few approaches that attempted to use both structural information,
including MGOMP \cite{lozanoblock} and ``group sparsity for multi-task learning''
\cite{tseng2009coordinate}.
MGOMP proposed to select the groups of regression coefficients 
from a predefined set of grouped variables in a greedy fashion, and 
\cite{tseng2009coordinate} proposed to find the groups of inputs that 
influence the group of outputs.
However, both methods may have limits on the number or the shapes of sparsity patterns that
can be induced in $\bB$. 
For example, given a large number of input groups $|\mathcal{G}|$ and output groups $|\mathcal{H}|$
(e.g. $|\mathcal{G}|>10^5, \; |\mathcal{H}|>10^3$ for genome data)
the scalability of MGOMP can be substantially affected since it needs to select
the groups of coefficients from all possible combinations of input and output groups. 
For \cite{tseng2009coordinate}, only disjoint block sparsity patterns are considered, hence
it may not capture the sparsity patterns where the grouped variables overlap.

In this paper, we address
the problem of exploiting both the input and output structures in a high-dimensional 
linear regression setting practically encountered in eQTL mapping. 
Furthermore, to detect epistatic (i.e., interaction) effects between SNP pairs,
we additionally expand the input space to include pairwise terms (i.e., $x_i x_j$'s) guided by biological information, which necessitates 
attentions for avoiding excessive input dimension that can make the problem computationally prohibitive.   
Our main contributions can be summarized as follows:
\begin{enumerate}
\item 
We propose a highly general regression model 
with structured input-output regularizers called
``jointly structured input-output lasso'' (SIOL)
that discovers structured associations between SNPs and expression traits (Section \ref{sec:methods}).  
\item 
We develop a simple and highly efficient optimization method called
``hierarchical group thresholding'' (HiGT)
for solving the proposed regression problem 
under complex sparsity-inducing penalties in a very high-dimensional space (Section \ref{sec:optimization}). 
\item Extending 
SIOL, we propose ``structured polynomial multi-task regression'' to efficiently model non-additive SNP-SNP interactions guided by 
genetic interaction networks 
(Section \ref{sec:model_interaction}). 
\end{enumerate}

Specifically, given knowledge of the groupings of the inputs (i.e., SNPs) and outputs (i.e., traits) in a high-dimensional multi-task regression setting, we employ an $L_1/L_2$ norm over such structures 
to impose a group-level sparsity-inducing penalty simultaneously over both the columns and the rows of the regression coefficient matrix $\bB$
(In our setting, a row corresponds to coefficients regressing all SNP (or SNP pair) inputs to a particular trait output, 
thus reflecting possible epistatic effects;
and a column corresponds to coefficients regressing a particular SNP (or SNP pair) input to all trait outputs in question, 
thus reflecting possible pleotropic effects). 
Given reliable input and output structures, rich structured sparsity 
can increase statistical power significantly since
it makes it possible to borrow information not only within different output or input variables, but also across output and input variables. 

The sparsity-inducing penalties on both the inputs and 
outputs in SIOL introduce a non-differentiable 
and non-separable objective in an extremely high-dimensional optimization space, 
which prevents standard optimization methods such as the interior point \cite{nesterov1987interior}, the coordinate-descent \cite{friedman2007pathwise}, or even the recently invented union of supports \cite{jacob2009group} algorithms to be directly applied. 
We propose a simple and efficient algorithm called ``hierarchical 
group-thresholding'' 
to optimize our regression model with complex structured regularizers.
Our method is an iterative optimization algorithm, designed to handle complex structured regularizers 
for very large scale problems.
It starts with a non-zero $\bB$ (e.g. initialized by ridge regression \cite{hoerl1970ridge}), 
and progressively discards irrelevant groups of covariates using thresholding operations.
In each iteration, we also update the coefficients of the remaining covariates. 
To speed up our method, we employ 
a directed acyclic graph (DAG) where nodes represent
the zero patterns encoded by our input-output structured regularizers
at different granularity, and edges indicate the inclusion relations
among them.
Guided by the DAG,
we could efficiently discard irrelevant covariates.

As our third contribution, 
we consider non-additive pairwise interaction effects between the input variables, 
in a way that avoids a quadratic blow-up of the input dimension. 
In eQTL mapping studies, it is not uncommon that the effect of one SNP 
on the expression-level of a gene is dependent on the genotype of another SNP, 
and this phenomenon is known as epistasis. 
To capture pairwise epistatic effects of SNPs on the trait variation, we 
additionally consider non-additive interactions between the input covariates. 
However, in a typical eQTL mapping, as
the input lies in a very high-dimensional space, it is computationally and statistically infeasible
to consider all possible input pairs. 
For example,
for $J$ inputs (e.g. $500,000$ for a typical genome data set), 
we have $O(J^2)$ candidate input pairs,
and learning with all of them will require a significantly large sample size.
Many of the previous approaches for learning the epistatic interactions relied
on pruning candidate pairs based on the observed data \cite{Rat:96} or constructing candidate pairs from
individual SNPs that were selected based on marginal effects in the previous learning phase without modeling interactions \cite{Taskar:10,devlin2003analysis}.
A main disadvantage of the later approach is that it will miss
pairwise interactions when they have no or little individual effects
on outputs. 
Instead of choosing candidate SNP pairs based on only marginal effects, 
we propose to use 
genetic interaction network \cite{costanzo2010genetic}
constructed from large-scale biological experiments 
to consider biologically plausible candidate pairs.

The rest of this paper is organized as follows. 
In Section 2, we discuss 
previous works on learning a sparse
regression model with prior knowledge on either output or input structure. 
In Section 3, we introduce our proposed model 
``jointly structured input-output lasso'' (SIOL). 
To solve our regression problem, we present an efficient optimization method 
called ``hierarchical group-thresholding'' (HiGT) in Section 4.
We further extend our model to consider pairwise interactions among input
variables and propose ``structured polynomial multi-task regression'' in Section 5.
We demonstrate the accuracy of recovered structured sparsity and the speed
of our optimization method in Section 6 via simulation study, and 
present eQTLs having marginal and interaction effects in yeast that we identified in Section 7.
A discussion is followed in Section 8.

\section{Background: Linear Regression with Structured Sparsity}
\label{sec:background}

In this section, we lay out the notation and then review existing sparse regression 
methods that recover a structured sparsity pattern in the estimated regression coefficients
given prior knowledge on input or output structure.

\subsection{Notation for matrix operations}

Given a matrix $\mathbf{B} \in \mathbb{R}^{K \times J}$, we denote
the $k$-th row by $\bm{\beta}_k$, the $j$-th column
by $\bm{\beta}^j$, and the $(k,j)$ element 
by $\beta_k^j$. 
$\lVert \cdot \rVert_F$ denotes
the matrix Frobenius norm, $\lVert \cdot \rVert_1$ denotes an $L_1$ norm 
(entry-wise matrix $L_1$ norm for a matrix argument), and $\lVert \cdot \rVert_2$
represents an $L_2$ norm. 
Given the set of {\it column} groups $\mathcal{G}=\{{\bg}_1, \ldots, {\bg}_{|\mathcal{G}|}\}$ 
defined as a subset of the power set of $\{1, \ldots, J\}$, 
$\bm{\beta}_k^{\bg}$ represents the row vector with
elements $\{\beta_k^j: j \in \bg, \bg \in \mathcal{G} \}$, which is
a subvector of $\bm{\beta}_k$ due to group $\bg$. 
Similarly, for the set of {\it row} groups $\mathcal{H}=\{{\bh}_1, \ldots, {\bh}_{|\mathcal{H}|}\}$ 
over $M$ rows of matrix $\bB$, 
we denote by $\bm{\beta}_{\bh}^{j}$ the column subvector with
elements $\{\beta_k^j: k \in \bh, \bh \in \mathcal{H} \}$. 
We also define the submatrix of ${\bB}_{\bh}^{\bg}$ as 
a $|\bh| \times |\bg|$ matrix with elements $\{\beta_k^j: k \in \bh, \; j \in \bg, \;
\bh \in \mathcal{H}, \; \bg \in \mathcal{G} \}$. 

\subsection{Sparse estimation of linear regression}

Let $\bX \in \mathbb{R}^{J \times N}$ be the input data for  $J$ inputs and $N$ individuals, 
and $\bY \in \mathbb{R}^{K \times N}$ be the output data 
for $K$ outputs. We model the functional mapping from the common
$J$-dimensional input space to the $K$-dimensional output space, using a linear
model parametrized by unknown regression coefficients $\bB \in \mathbb{R}^{K \times J}$
as follows:
\begin{eqnarray}
\bY = \bB \bX + \bE, \nonumber
\label{eq:reg}
\end{eqnarray}
where $\bE \in \mathbb{R}^{K \times N}$ is a matrix of noise terms whose elements
are assumed to be identically and independently distributed as Gaussian with zero mean 
and the identity covariance matrix.
Throughout the paper, we assume that $x_j^{i}$'s and $y_k^i$'s 
are standardized
such that all rows of $\bX$ and $\bY$ have zero mean and a constant variance, 
and consider a model without an intercept.
In eQTL analysis, inputs are genotypes for $J$ loci encoded as 0, 1, or 2 in terms of
the number of minor alleles at a given locus, and output data are given as
expression levels of genes measured in a microarray experiment.
Then, the regression coefficients represent the strengths of associations 
between genetic variations and gene expression levels.

Our proposed method for estimating the coefficients $\bB$ is based on
a 
group-structured multi-task regression approach that extends 
existing regularized regression approaches including lasso \cite{tibshirani1996regression}, 
group lasso \cite{yuan2006model}  and multi-task lasso \cite{obozinski2006multi}, 
which we briefly review below in the context of our eQTL mapping problem. 
When $J >> N$ and only a small number of inputs are expected
to influence outputs, lasso has been widely used and shown effective in
selecting the input variables relevant to outputs and setting the elements of 
$\bB$ for irrelevant inputs to zero \cite{zhang2008sparsity}.
Lasso obtains a sparse estimate of regression coefficients by optimizing 
the least squared error criterion with an $L_1$ penalty over $\bB$ as follows:
\begin{eqnarray}
\min_{\bB} \frac{1}{2}
\lVert \bY - \bB \bX  \rVert_F^2 
 + \lambda \lVert \bB \rVert_1,
\label{eq:lasso}
\end{eqnarray}
where $\lambda$ is the tuning parameter that determines the amount of 
penalization. 
The optimal value of $\lambda$ 
can be determined by cross validation or via an information-theoretic test based on BIC. 
As in eQTL analysis it is often believed that the expression level of each gene is affected by
a relatively small number of genetic variations in the whole genome, lasso provides
an effective tool for identifying eQTLs from a large number of genetic variations.
Lasso has been previously applied to eQTL analysis \cite{brown2011application} 
and more general genetic association
mapping problems \cite{wu2009genome}.

While lasso considers the input variables independently to select relevant
inputs with non-zero regression coefficients, we may have prior knowledge 
on how related input variables are grouped together and want to perform variable
selection at the group level rather than at the level of individual inputs.
Grouped variable selection approach can combine the statistical strengths across
multiple related input variables to achieve higher power for detecting
relevant inputs in the case of low signal-to-noise ratio.
Assuming the grouping structure over inputs are available as 
$\mathcal{G} = \{{\bg}_1, \ldots, {\bg}_{|\mathcal{G}|}\}$, which is a subset
of the power set of $\{1, \ldots, J\}$, group lasso uses $L_1/L_2$ penalization
to enforce that all of the members in each group of input variables are jointly 
relevant or irrelevant to each output. Group lasso obtains an estimate
of $\bB$ by solving the following optimization problem:
\begin{eqnarray}
\min_{\bB} \frac{1}{2} \lVert \bY - \bB \bX  \rVert_F^2 
+ \lambda \sum_{k=1}^K \sum_{\bg \in \mathcal{G}} {\lVert \bm{\beta}^{\bg}_{k} \rVert_2}  
\label{eq:lasso_g},
\end{eqnarray}
where $\lambda$ is the tuning parameter. The second term in the above
equation represents an $L_1/L_2$ penalty over each row $\bm{\beta}_k$ of $\bB$ for 
the $k$-th output given $\mathcal{G}$, defined by 
$\lVert \bm{\beta}_k \rVert_{L_1/L_2} = \sum_{\bg \in \mathcal{G}} \lVert \bm{\beta}_k^{\bg} \rVert_2$.
The $L_2$ part of the penalty plays the role of enforcing a joint selection
of inputs within each group, whereas the $L_1$ part of the penalty is applied
across different groups to encourage a group-level sparsity.
Group lasso can be applied to an eQTL mapping problem given biologically
meaningful groups of genetic variations that are functionally related.
For example, rather than individual genetic variations acting
independently to affect (or not affect) gene expressions, the variations are 
often related through pathways that consist of multiple genes participating
in a common function. Thus, genetic variations can be grouped according to pathways 
that contain genes carrying those genetic variations. Then, given this grouping, 
group lasso can be used to select groups of genetic variations
in the same pathways as factors influencing gene expression levels \cite{silverpathway}.

Instead of having groups over inputs with outputs being independent as in group lasso, 
the idea of using $L_1/L_2$ penalty for grouped variable 
selection has also been applied to take advantage of the relatedness among outputs in multiple
output regression.
In multi-task regression for union support recovery \cite{obozinski2006multi}, 
one assumes that all the outputs share a common support of relevant input variables 
and try to recover shared sparsity patterns across multiple outputs 
by solving the following optimization problem:
\begin{eqnarray}
\min_{\bB} \frac{1}{2}
\lVert \bY - \bB \bX  \rVert_F^2 
+ \lambda \sum_{j=1}^J \sum_{\bh \in \mathcal{H}} \lVert \bm{\beta}_{\bh}^j \rVert_2,  \label{eq:multi_task2} 
\end{eqnarray}
where 
$\lambda$ can be determined by cross-validation.
In eQTL mapping, as gene expression levels are often correlated for the genes that participate
in a common function, it is reasonable to assume that those coexpressed genes 
may be influenced by common genetic variations.
If gene module information is available, one can use the above model to 
detect genetic variations influencing the expressions
of a subset of genes within each gene module. 
This strategy corresponds to a variation of the standard group lasso, where group is defined 
over outputs rather than inputs.

Extending the idea of lasso and group lasso, we may have 
group and individual level sparsity simultaneously using 
combined $L_1$ and $L_1/L_2$ penalty.
In group lasso, if a group of coefficients is not jointly set to zero, all the members 
in the group should have non-zero values.
However, sometimes it is desirable to set some members of the
group to zero if they are irrelevant to outputs.
Sparse group lasso \cite{friedman2010note} is proposed to address 
the cases where 
groups of coefficients 
include both relevant and irrelevant 
ones.
Using convex combination of $L_1$ and $L_1/L_2$ norms,
it solves the following convex optimization problem:
\begin{eqnarray}
\min_{\bB} \frac{1}{2}
\lVert \bY - \bB  \bX \rVert_F^2 
+ \lambda_1  \lVert \bB \rVert_1
+ \lambda_2  \sum_{k=1}^K \sum_{\bg \in \mathcal{G}} {\lVert \bm{\beta}^{\bg}_{k} \rVert_2},
\label{eq:sparse_group_lasso} 
\end{eqnarray}
where $\lambda_1$ and $\lambda_2$ determine the individual and
group level sparsity, respectively.
The $L_1/L_2$ penalty shrinks
groups of coefficients to zero, and at the same time,
$L_1$ penalty  sets irrelevant coefficients to zero individually within each group.

Our proposed model is motivated by 
group lasso, multi-task lasso
and sparse group lasso, each of which can
exploit pre-defined groupingness of input 
or output variables to  
achieve better statistical power.
In the next section, 
we will extend the existing models in such a way
that we can use
the groups in both input and output spaces simultaneously.
Adopting the idea of sparse group lasso, 
we will also support variable selection
at individual levels.

\section{Jointly Structured Input-Output Lasso} 
\label{sec:methods}

In this section, we propose SIOL 
that incorporates structural constraints on both the inputs and outputs.
The model combines the mixed-norm regularizers for the groups of 
inputs and outputs, which leads to the following optimization problem:
\begin{subequations}
\label{equ:reg5}
\begin{align}
\min \frac{1}{2}
\lVert \bY - \bB \bX  \rVert_F^2 
&  + \lambda_1   
\lVert \bB \rVert_1,
\label{equ:reg5_b}     \\
  &  + \lambda_2  
\sum_{k=1}^K \sum_{{\bg} \in \mathcal{G}} {\lVert \bm{\beta}^{\bg}_{k} \rVert_2}    
\label{equ:reg5_c}   \\
  & + \lambda_3 
\sum_{j=1}^J \sum_{{\bh} \in \mathcal{H}} \lVert \bm{\beta}_{\bh}^j \rVert_2,  
\label{equ:reg5_d}
\end{align}
\end{subequations}
where Eq. (\ref{equ:reg5_c})
incorporates the  groups of inputs $\mathcal{G}=\{{\bg}_1, \ldots, {\bg}_{|\mathcal{G}|}\}$,
Eq. (\ref{equ:reg5_d}) 
incorporates the groups of the outputs 
$\mathcal{H}=\{{\bh}_1, \ldots, {\bh}_{|\mathcal{H}|}\}$, 
and Eq. (\ref{equ:reg5_b}) allows us to select individual coefficients.
Note that it is possible that there are overlaps between 
$\bm{\beta}_k^{\bg}$ and $\bm{\beta}_k^{\bg'}$, 
between $\bm{\beta}_{\bh}^j$ and $\bm{\beta}_{\bh'}^j$, and
between $\bm{\beta}_k^{\bg}$ and $\bm{\beta}_{\bh}^j$, where 
$\bg \neq \bg', \bh \neq \bh'$ and
$\bg, \bg' \in \mathcal{G}, \; \bh, \bh' \in \mathcal{H}$. 
The overlaps make it challenging to optimize Eq. (\ref{equ:reg5}), and 
this issue will be addressed by our optimization method in Section \ref{sec:optimization}.

Let us characterize the structural constraints imposed by the penalties in our model. 
In our analysis, we investigate a block of coefficients involved in one output group $\bh$ and 
one input group $\bg$, i.e, $\bB_{\bh}^{\bg}$.
We start with Karush-Kuhn-Tucker (KKT)
condition for Eq. (\ref{equ:reg5}): 
\begin{equation}
(\mathbf{y}_k-\bm{\beta}_k \bX )(\mathbf{x}_j)^T = \lambda_1 s_k^j + \lambda_2 c_k^j + \lambda_3 d_k^j,
\label{eq:subgradients}
\end{equation}
where $s_k^j$, $c_k^j$, and $d_k^j$ are the subgradient of
Eq. (\ref{equ:reg5_b}), Eq. (\ref{equ:reg5_c}), and Eq. (\ref{equ:reg5_d})
with respect to $\beta_k^j$, respectively.
For simple notation, we also define 
${\br}_k^j=\mathbf{y}_k-\sum_{l\neq j} \beta_k^l \bx_l$. 

First, we consider the case where all coefficients in $\bB_{\bh}^{\bg}$ become
zero simultaneously, i.e., $\bB_{\bh}^{\bg} = \bm{0}$.
Using KKT condition in Eq. (\ref{eq:subgradients}), we
can see that $\bB_{\bh}^{\bg} = \bm{0}$ if and only if
\begin{align}
\label{eq:sub1}
\sum_{k \in \bh}\sum_{j \in \bg}  \left\{ \br_k^j(\bx_j)^T  - \lambda_1 s_k^j\right\}^2 
\leq \sum_{k \in \bh}\sum_{j \in \bg} \left( \lambda_2 c_k^j + \lambda_3 d_k^j \right)^2
\leq \left( \lambda_2 \sqrt{|\bh|} + \lambda_3 \sqrt{|\bg|} \right)^2.
\end{align}
This condition is due to Cauchy-Schwarz inequality,
$\sum_{j\in \bg}(c_k^j)^2 \leq 1$, and $\sum_{k\in \bh}(d_k^j)^2 \leq 1$. 
Here if $\lambda_1$, $\lambda_2$ and $\lambda_3$ are large, 
$\bB_{\bh}^{\bg}$ is likely to be zero jointly. 
This structural sparsity is useful to filter out a large number of
irrelevant covariates since it considers both the group of
correlated inputs $\bg$ and the group of correlated outputs $\bh$ simultaneously.

Our model also inherits grouping effects for only input (or output) groups.
For the analysis of such grouping effects, we fix the groups of zero coefficients that overlap with, say, an input group $\bm{\beta}_k^{\bg}$.
Formally speaking, let us define $\bm{\xi} = \{j: (\bm{\beta}_{\bh'}^j = \bm{0}, j \in \bg, \bh' \in \mathcal{H})
\lor (\bm{\beta}_{k}^{\bg'} = \bm{0}, j \in \bg' \land \bg) \}$,
and fix $\beta_k^j$s for all $j \in \bm{\xi}$.
Using the KKT condition in Eq. (\ref{eq:sub1}), $\bm{\beta}_k^{\bg} = \bm{0}$ if 
\begin{align}
\sum_{j \in \bg - \bm{\xi}}  \left\{ {\br}_k^j(\bx_j)^T  - \lambda_1 s_k^j\right\}^2
\leq \sum_{j \in \bg - \bm{\xi}} \left( \lambda_2 c_k^j + \lambda_3 d_k^j \right)^2
\leq \lambda_2^2. 
\label{eq:sub2}
\end{align}
Here, we know that $d_k^j = 0$ for $j \in \bg - \bm{\xi}$ ($\beta_k^j = 0$ and $\bm{\beta}_{\bh}^j \neq \bm{0}$) and 
$\lambda_2 \sum_{j \in \bg} (\beta_k^j)^2 = \lambda_2 \sum_{j \in \bg - \bm{\xi}} (\beta_k^j)^2$,
and hence  $\sum_{j \in \bg - \bm{\xi}} \left( \lambda_2 c_k^j + \lambda_3 d_k^j\right)^2 \leq \lambda_2^2$.
This technique was previously introduced in \cite{yuanefficient} to handle overlapping group lasso.
One can see that if the size of $\bm{\xi}$ is large, 
$\bm{\beta}_{k}^{\bg}$ tends to be zero together
since it reduces the left-hand side of Eq. (\ref{eq:sub2}).
This behavior explains the correlation effects between input and output group structures. 
When a group of coefficients ($\bm{\beta}_k^{\bg}$, $\bm{\beta}_{\bh}^{j}$) corresponding to an input group or an output group become zero, they
affect other groups of coefficients that overlap with them; and 
the overlapped coefficients are more likely to be zero.
These correlation effects between overlapping groups are desirable for inducing 
appropriate structured sparsity as it allows us to share information across different 
inputs and different outputs simultaneously.
We skip the analysis of the grouping effects for output groups as
the argument is the same 
except that the input and output group are reversed.

Finally, we also have individual sparsity due to $L_1$ penalty in Eq. (\ref{equ:reg5_b}).
In this case, let us assume that 
$\bm{\beta}_{k}^{\bg} \neq \bm{0}$ and $\bm{\beta}_{\bh}^{j} \neq \bm{0}$
since if the group of coefficients is zero, we automatically have $\beta_k^j=0$.
Using the KKT condition, $\beta_k^j=0$ if and only if
\begin{align}
|{\br}_k^j(\mathbf{x}_j)^T |
\leq \lambda_1.
\label{eq:sub3}
\end{align}
It is equivalent to the condition of lasso that sets a regression coefficient to zero. 
Note that if $\lambda_2=\lambda_3 = 0$, 
we have sparsity only at the individual levels, and our model is the same as lasso.
When a group of coefficients contains
both relevant and irrelevant ones, 
we can set the irrelevant coefficients to zero using Eq. (\ref{eq:sub3}).

We briefly mention the three tuning parameters ($\lambda_1$, $\lambda_2$, $\lambda_3$) 
which can be determined by cross validation. 
It is often computationally expensive to search for optimal parameters in 3-dimensional grid.
In practice, instead, we use the following tuning parameters:
$\lambda_2 = \lambda_2'\lambda_3'$  and 
$\lambda_3 = (1-\lambda_2')\lambda_3'$.
Here $\lambda_2$ configures the mixing proportion of
input and output group structures, and 
$\lambda_3$ is the scaling factor that determines
the degree of penalization for the input and output groups.
In this setting, we also have three regularization parameters, however,
it helps us to reduce the search space of the 
tuning parameters as we know the range of $\lambda_2'$ ($0\leq \lambda_2' \leq 1$). 

Let us discuss the statistical and biological benefits of
our model.
First, our model can capture rich structured sparsity in $\bB$.
The structured sparsity patterns include zero (sub)rows, zero (sub)columns and
zero blocks of $\bB_{\bh}^{\bg}$. 
It is impossible to have such rich sparsity patterns if we
use one part of information on either input or output side. 
For example, group lasso \cite{yuan2006model} 
or multi-task lasso \cite{obozinski2006multi} consider 
structured sparsity patterns in either rows or columns in $\bB$. 
Second, our model is robust to the 
groups which 
contain both relevant and irrelevant coefficients.
If predefined groups of inputs and outputs
are unreliable, our model may still work 
since the irrelevant coefficients can be set to zero individually via $L_1$ penalty 
even when their groups are not jointly set to zero.
Third, the grouping effects induced by our model in Eq. (\ref{eq:sub1}, \ref{eq:sub2})
show that we can use 
the correlation effects between input and output groups.
When we have reliable input and output groups,
the advantage from the structural information will be
further enhanced by the correlation effects in addition to
the sum of the benefits of both input and output groups.

When applied to GWA mapping of eQTLs, our model offers a number of desirable properties.
It is likely that our model can detect association SNPs with low signal-to-noise ratio by 
taking advantage of rich structural information.
In GWA studies, one of the main challenges 
is to detect SNPs having weak signals with limited sample size.
In complex diseases such as cancer and diabetes, biologists 
believe that multiple SNPs are jointly responsible for diseases but not necessarily with
strong marginal effects \cite{mccarthy2008genome}. 
However,
such causal SNPs are hard to detect mainly due to insufficient number of samples.
Our model can deal with this challenge by taking advantage of both input and output group
structures.
First, by grouping inputs (or SNPs), we can increase the signal-to-noise ratio.
Suppose each SNP has small signal marginally, 
if a group of coefficients is relevant,
their joint strength will be increased, and it is unlikely that they are jointly set to zero.
On the other hand, if a group of coefficients is irrelevant, 
their joint strength will still be small, and it is likely that they are set to zero.
Second, taking advantage of the output groups, we can share information across 
the correlated outputs, 
and it decreases the sample size required for successful support recovery 
\cite{negahban2011simultaneous}.
Overall, to detect causal SNPs having small effects,
our model increases signal-to-noise ratio by grouping 
the SNPs, and simultaneously
decreases the required number of samples by grouping
phenotypic traits.

Unfortunately, the optimization problem resultant from Eq.  (\ref{equ:reg5}) is non-trivial.
One may find out that each $\beta_k^j$ appears 
in all the three penalties of Eq. (\ref{equ:reg5_b} -- \ref{equ:reg5_d}). 
Thus, our structured regularizer is non-separable, which makes 
simple coordinate descent optimization inapplicable. 
The overlaps between/within input and output groups
add another difficulty.
Furthermore,
we must induce appropriate sparsity patterns (i.e., exact zeros)
in addition to the minimization of Eq.  (\ref{equ:reg5}), 
therefore approximate methods based on merely relaxing the shrinkage functions are not appropriate.
In the following section, we propose ``hierarchical group thresholding'' method (HiGT)
that efficiently solves our optimization problem
with hierarchically organized thresholding operations.

\section{Optimization method}
\label{sec:optimization}

\begin{figure} 
\centering
\subfigure[]{\includegraphics[width=0.8\textwidth]{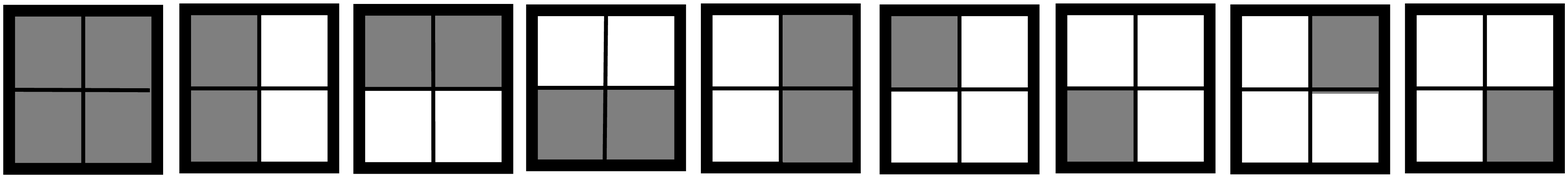}}
\subfigure[]{\includegraphics[width=0.35\textwidth]{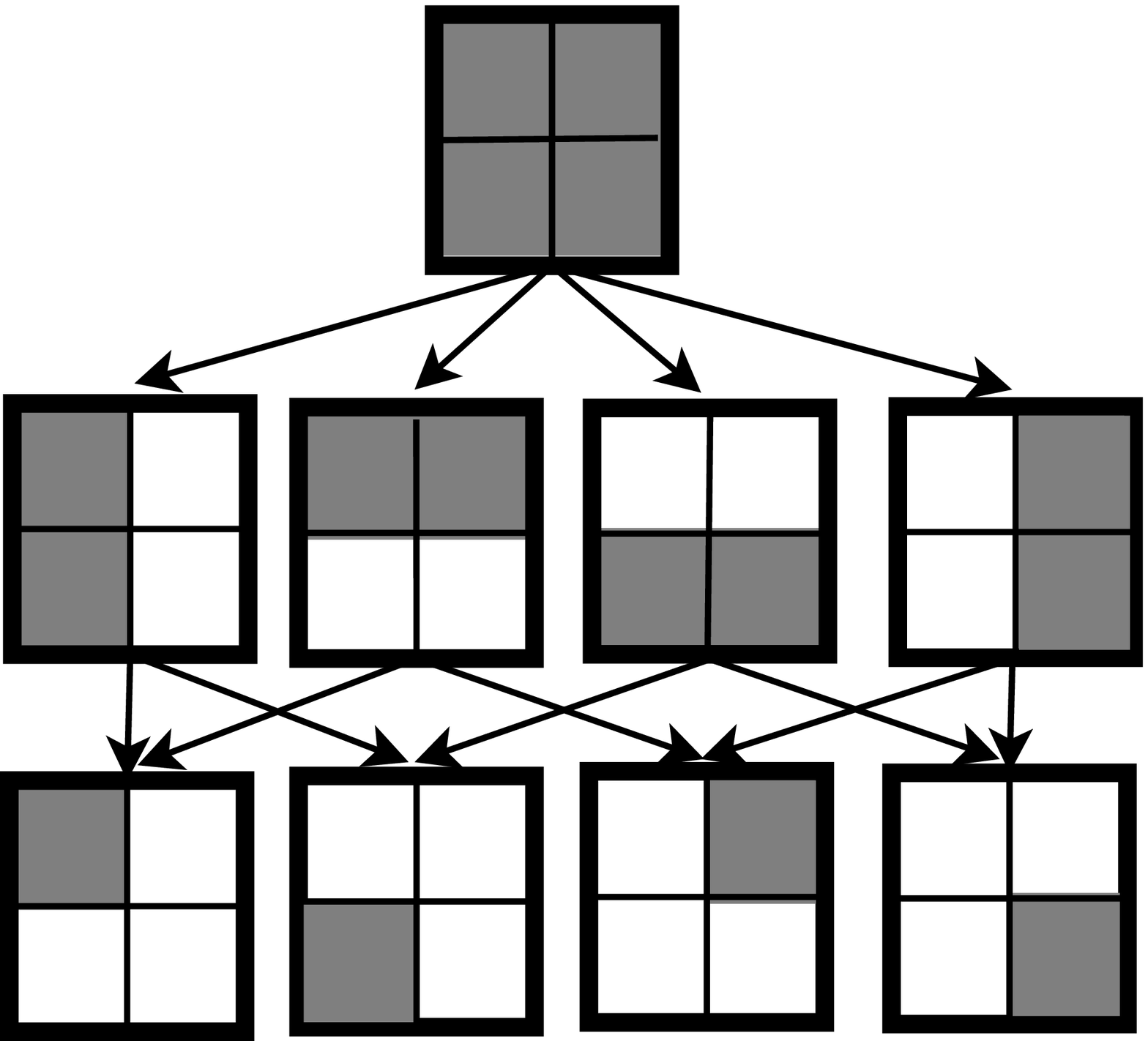}}
\caption{
Sparsity patterns of $\bB_{\bh}^{\bg}$ and a DAG constructed with the sparsity patterns.
The shaded area shows zero entries.
(a) All possible zero patterns of $\bB_{\bh}^{\bg}$ 
that can be induced by Eq. (\ref{equ:reg5_b},\ref{equ:reg5_c},\ref{equ:reg5_d}) when 
$\bg = \{1,2\}$ and $\bh = \{1,2\}$.
(b) An example of a DAG that 
contains the zero patterns of $\bB_{\bh}^{\bg}$.
The root node contains zero pattern for $\bB_{\bh}^{\bg}=\bm{0}$,
and the internal nodes represent the zero patterns for $\bm{\beta}_{\bh}^{j}=\bm{0}$ (one column is zero) 
or $\bm{\beta}_{k}^{\bg}=\bm{0}$ (one row is zero). The leaf nodes denote $\beta_{k}^{j}=0$.
In the DAG, the zero pattern of children nodes should be a subset of 
their parent nodes' zero patterns.
}
\label{fig:all_zeros}
\end{figure}

In this section, we propose our method to optimize Eq. (\ref{equ:reg5}).
We start with a non-zero $\bB$ initialized by other methods (e.g. ridge regression), 
and always reduce the set of non-zero $\beta_k^j$s using thresholding operations 
as our procedure proceeds.
Our framework is an iterative procedure consisting 
of two steps. First, we set the groups (or individual) of regression coefficients
to zero by checking optimality conditions (called thresholding) 
as we walk through a predefined 
directed acyclic graph (DAG). 
When we walk though the nodes in the DAG, 
some $\beta_k^j$s might not achieve zero. 
Second, we update only these non-zero $\beta_k^j$s using any available optimization techniques.

Let us first 
characterize the zero patterns induced by Eq. (\ref{equ:reg5_b} -- \ref{equ:reg5_d}).
We separately consider a block of $\bB$ which consists of 
one input group $(\bg\in \mathcal{G})$ and one output group $(\bh\in \mathcal{H})$.
Our observation tells us that there are grouping effects 
(to be zero simultaneously) for each $\bg$ and $\bh$: 
$\bm{\beta}_k^{\bg}=\bm{0}$ 
and $\bm{\beta}_{\bh}^{j}=\bm{0}$. 
We also have $\bB_{{\bh}}^{{\bg}} = \bm{0}$
when $\bm{\beta}_k^{\bg} = \bm{0}, \; \forall k\in \bh$
or $\bm{\beta}_{\bh}^{j} = \bm{0}, \; \forall j \in \bg$. 
Each covariate can also be zero, i.e, $\beta_k^j = 0$
due to the $\ell_1$ penalty in Eq. (\ref{equ:reg5_b}).
Figure \ref{fig:all_zeros}(a) shows 
all the possible
zero patterns of $\bB_{\bh}^{\bg}$ induced 
by Eq. (\ref{equ:reg5_b} -- \ref{equ:reg5_d}). 
Given these sparsity patterns, 
to induce structured sparsity, 
one might be able to check whether or not
these zero patterns satisfy optimality conditions 
and discard irrelevant covariates accordingly. 
However, this approach may be inefficient as it 
needs to examine the large number of 
zero patterns. 
Instead, to efficiently check the zero patterns,
we will construct a DAG, and exploit the inclusion relationships
between the zero patterns.
The main idea is that we want to be able to check all zero patterns 
by traversing the DAG while avoiding unnecessary optimality checks.

In Figure \ref{fig:all_zeros}(b), we show an example of the DAG for
$\bB_{\bh}^{\bg}$ when $\bg = \{1,2\}$  
and $\bh = \{1,2\}$. 
We denote the set of all possible zero patterns of $\bB$ 
by $\mathcal{Z} = \{Z_1,\ldots,Z_{|\mathcal{Z}|}\}$. 
For example, $Z_1$ can be a zero pattern for 
$\bB_{\bh}^{\bg} = \bm{0}$ (the root node in Figure \ref{fig:all_zeros}(b)).
Let us denote $\bB(Z_t)$ by the coefficients of $\bB$ corresponding to $Z_t$'s zero pattern.
Then we define the DAG as follows:
A node is represented by $Z \in \mathcal{Z}$, 
and there exists a directed edge from 
$Z_1 \in \mathcal{Z}$ to $Z_2 \in \mathcal{Z}$ if and only if $Z_1 \supset Z_2$
and $\nexists Z \in \mathcal{Z}: Z_1 \supset Z \supset Z_2$.
For example, in Figure \ref{fig:all_zeros}(b), the zero patterns
of the nodes in the second level include the zero patterns of their children.
In general, when we have multiple input and output groups,
we can generate a DAG for each $\bB_{{\bh}}^{{\bg}}$ separately 
and then connect all the DAGs to the root node for $\bB = \bm{0}$.
This graph is originated from Hasse diagram \cite{cameron1994combinatorics}
and it was previously utilized for finding a minimal set of 
groups for inducing structured sparsity \cite{jenatton2009structured}. 

We can readily observe that our procedure has the following properties:
\vspace{-0.2cm}
\begin{itemize}
\item Walking through the DAG, we can check all possible zero patterns
explicitly without resorting to heuristics or approximations.
\item If $\bB(Z) = \bm{0}, \; Z \in \mathcal{Z}$, we know that
all the descendants of $Z$ are also zero due to
the inclusion relations of the DAG. Hence, we can ``skip''
to check the optimality conditions that the descendants of 
$Z$ are zero.  
\end{itemize}
\vspace{-0.2cm}

Considering these properties,
we develop our optimization framework for the following reasons.
First, we can achieve accurate zero patterns in $\bB$
since we check all possible sparsity patterns when walking through the DAG. 
Second, if $\bB$ is sparse,
our framework is very efficient since 
we can skip the optimality checks for many zero patterns in $\mathcal{Z}$. 
Mostly we will check only nodes
located at the high levels of the DAG.
Third, our framework is simple to implement. All we need
is to check whether each node in the DAG attains zero and 
update non-zero $\beta_k^j$s only when necessary.

Specifically, our hierarchical group-thresholding employs the following procedure:

\vspace{-0.1cm}
\begin{enumerate}
\vspace{-0.1cm}
\item Initialize a non-zero $\bB$ using any available methods (e.g. ridge regression).
\vspace{-0.1cm}
\item Construct a DAG that contains all zero patterns of $\bB$  
that can be induced by the penalty in Eq. (\ref{equ:reg5_b}, \ref{equ:reg5_c}, \ref{equ:reg5_d}).
\vspace{-0.1cm}
\item 
\label{algo:iter1}
Use depth-first-search (DFS) to traverse the DAG, and check the optimality conditions
to see if the zero patterns at each node $Z$ achieve zero.
If $\bB(Z) = 0$ 
or $Z$ satisfies the optimality condition to be zero, 
set $\bB(Z) = 0$,
skip the descendants of $Z$, and visit the next node according to the DFS order.
\vspace{-0.1cm}
\item 
\label{algo:iter2}
For those of $\beta_k^j$'s which did not achieve zero in the previous step, 
update the coefficients of the non-zero $\beta_k^j$'s using 
any available optimization algorithms.
\vspace{-0.1cm}
\item Iterate step \ref{algo:iter1} and \ref{algo:iter2} until Eq. (\ref{equ:reg5}) converges.
\end{enumerate}
\vspace{-0.1cm}

Bellow we briefly present the derivations of the three
ingredients of our optimization framework that include
1) the construction of a DAG, 
2) the optimality condition of each $Z \in \mathcal{Z}$ in the DAG and 
3) the rule for updating non-zero regression coefficients.
Our optimization method is summarized in Algorithm \ref{alg:hGroupThres}.

\begin{algorithm}[!ht]
\caption{Hierarchical group-thresholding method for Eq. (\ref{equ:reg5})}
{\scriptsize
\begin{algorithmic}
\label{alg:hGroupThres}
\STATE $\bB \leftarrow \mbox{coefficients estimated by ridge regression}$
\STATE $\mathcal{G} \leftarrow \mbox{groups of inputs}$
\STATE $\mathcal{H} \leftarrow \mbox{groups of outputs}$
\STATE $D(\mathcal{Z},\mathcal{E}) \leftarrow \mbox{DAG including all zero patterns}$
\STATE $\{Z_{(1)},Z_{(2)},\ldots,Z_{(|\mathcal{Z}|)}\}  \leftarrow \mbox{DFS order of $\mathcal{Z}$ in $D$}$
\REPEAT 
\STATE $t \leftarrow 1$
\WHILE{$t\leq |\mathcal{Z}|$}  
\IF{$Z_{(t)}$ contains $\bB_{\bh}^{\bg}=\bm{0}$}  
\STATE c $\leftarrow$ Eq. (\ref{equ:rule1})
\ELSIF{$Z_{(t)}$ contains $\bm{\beta}_k^{\bg}=\bm{0}$} 
\STATE c $\leftarrow$ Eq. (\ref{equ:rule2})
\ELSIF{$Z_{(t)}$ contains $\bm{\beta}_{\bh}^{j}=\bm{0}$} 
\STATE c $\leftarrow$ Eq. (\ref{equ:rule3})
\ELSIF{$Z_{(t)}$ contains $\beta_{k}^{j}=0$} 
\STATE c $\leftarrow$ Eq. (\ref{equ:rule4})
\ENDIF
\IF{c holds (condition for $\bB(Z_{(t)}) = \bm{0}$) or $\bB(Z_{(t)}) = \bm{0}$}
\STATE $\bB(Z_{(t)}) = \bm{0}$ (Set zero to $Z_{(t)}$'s zero pattern)
\STATE $t \leftarrow $ DFS order of $t'$ such that 
$Z_{(t')}$ is not a descendant of $Z_{(t)}$, $t'>t$
and $\nexists t{''}: t'>t{''}>t$
(Skip the descendants of $Z_{(t)}$)
\ELSIF{c $=$ Eq. (\ref{equ:rule4})}
\STATE Update $\beta_k^j$ using Eq. (\ref{equ:update_rule})
(Updating non-zero regression coefficients)
\STATE $t \leftarrow t+1$
\ELSE
\STATE $t \leftarrow t+1$
\ENDIF
\ENDWHILE
\UNTIL{convergence}
\end{algorithmic}
}
\end{algorithm}
\vspace{-0.3cm}

\paragraph{Construction of the DAG}
To generate the DAG, first we define the set of nodes
$\mathcal{Z}$. For each block of $\bB_{\bh}^{\bg}$, 
we are interested in the four types of zero patterns as follows:

\vspace{-0.4cm}
\begin{enumerate}
\item $\bB_{{\bh}}^{{\bg}}$ is zero: $\bB_{{\bh}}^{{\bg}} = \bm{0}$.
\item One row in $\bB_{{\bh}}^{{\bg}}$ is zero:
$\bm{\beta}_{k}^{{\bg}} = \bm{0}$, $k\in {\bh}$.
\item One column in $\bB_{{\bh}}^{{\bg}}$ is zero:
$\bm{\beta}_{{\bh}}^{j} = \bm{0}$, $j\in {\bg}$.
\item One regression coefficient in $\bB_{{\bh}}^{{\bg}}$ is zero:
$\beta_{k}^{j} = 0$, $k\in {\bh}$ and $j\in {\bg}$.
\end{enumerate}
\vspace{-0.3cm}

These zero patterns of 
$\bB$ are shown in Figure \ref{fig:all_zeros}(b)
when $|\bg| = |\bh| = 2$.
For example, Case 2 and 3 correspond to the nodes at the second level 
of the DAG.
For all $\bg \in \mathcal{G}$ and $\bh \in \mathcal{H}$, 
we can define nodes $Z \in \mathcal{Z}$ using the above zero patterns.
Then we need to determine the edges of the DAG by 
investigating the relations of the nodes.
We can also easily see that there exists the relationship among the zero patterns:
$\mbox{Case } 1 \supset \mbox{Case } 2, \mbox{Case } 3  \supset \mbox{Case } 4$.
Given the zero patterns and their relations, 
we create a directed
edge $Z_1 \rightarrow Z_2$ 
if and only if $Z_1 \supset Z_2$
and $\nexists Z \in \mathcal{Z}: Z_1 \supset Z \supset Z_2$.
In Figure \ref{fig:all_zeros}(b) we show an example of the DAG. 
Finally, we make a dummy root node and generate an edge from the
dummy node to the root of all DAGs for $\bB_{{\bh}}^{{\bg}}= \bm{0}$.

\paragraph{Optimality conditions for structured sparsity patterns}
Given a block of $\bB_{\bh}^{\bg}$, 
here we show optimality conditions for the four sparsity patterns:
(1) $\bB_{\bh}^{\bg} = \bm{0}$,
(2) $\bm{\beta}_k^{\bg} = \bm{0}$,
(3) $\bm{\beta}_{\bh}^j = \bm{0}$, and
(4) $\beta_k^j = 0$, 
($j \in \bg, k \in \bh, \bg \in \mathcal{G}, \bh \in \mathcal{H}$).
In Figure \ref{fig:all_zeros}(b), the root node
corresponds to the first case, the nodes at the second level 
correspond to the second and third case, and the leaf nodes correspond to
the fourth case.
Our derivation of the following optimality conditions 
use the fact that all zero coefficients are fixed, as
it makes it simple to deal with overlapping groups.
We denote the column and row indices of zero entries by 
$\bm{\eta} = \{j : \beta_{k}^j = 0, \; \forall j \in \bg, \; \forall k \in \bh\}$  and
$\bm{\gamma} = \{k : \beta_{k}^j = 0, \; \forall j \in \bg, \; \forall k \in \bh\}$.

First, the optimality condition for the first case is as follows: $\bB_{\bh}^{\bg} = \bm{0}$ if
\begin{align}
\label{equ:rule1}
\sum_{k \in \bh - \bm{\gamma}} \sum_{j \in \bg - \bm{\eta}} 
\left\{ {\br}_k^j (\mathbf{x}_j)^T  - \lambda_1 s_k^j \right\}^2
\leq \left(\lambda_2 \sqrt{|\bh|} 
+ \lambda_3 \sqrt{|\bg|}\right)^2, 
\end{align}
where
\[s_k^j = \left\{ 
\begin{array}{l l}
  \frac{{\br}_k^j (\mathbf{x}_j)^T  }{\lambda_1} 
  & \mbox{if $\left|\frac{ {\br}_k^j (\mathbf{x}_j)^T}{\lambda_1}\right| \leq 1$}\\ 
	sign\left(\frac{ {\br}_k^j (\mathbf{x}_j)^T}{\lambda_1}\right) 
  & \mbox{if $\left|\frac{ {\br}_k^j (\mathbf{x}_j)^T}{\lambda_1}\right| > 1$}.\\  
  \end{array} \right.\]
It is derived using KKT condition in Eq. (\ref{eq:subgradients})
and Cauchy-Schwarz inequality.
  
The second case of structured sparsity, i.e, $\bm{\beta}_k^{\bg}=\bm{0}$ is achieved if
\begin{align}
\label{equ:rule2}
\sum_{j \in {\bg - \bm{\eta}}} \left\{{\br}_k^j(\mathbf{x}_j)^T -\lambda_1 s_k^j \right\}^2
\leq  \lambda_2^2,  
\end{align}
and the optimality condition for the third case, i.e, $\bm{\beta}_{\bh}^{j}=\bm{0}$ is
\begin{align}
\label{equ:rule3}
\sum_{k \in \bh - \bm{\gamma}} \left\{{\br}_k^j(\mathbf{x}_j)^T - \lambda_1 s_k^j \right\}^2
\leq \lambda_3^2.
\end{align}
These conditions can be established using KKT condition in Eq. (\ref{eq:subgradients})
fixing all the zero coefficients.
Finally, assuming that 
$\bm{\beta}_{\bh}^{j} \neq \bm{0}$ and $\bm{\beta}_{k}^{\bg} \neq \bm{0}$,
the fourth case has the optimality condition of 
\begin{align}
\label{equ:rule4}
| {\br}_k^j (\mathbf{x}_j)^T| \leq \lambda_1.
\end{align}
 
\paragraph{Update rule for nonzero coefficients}
If all the above optimality conditions do not hold, we 
know that $\beta_k^j\neq 0$.
In this case, the gradient of Eq. (\ref{equ:reg5}) with respect to $\beta_k^j$ exists, and
we can update $\beta_k^j$ using any coordinate descent procedures.
With a little bit of algebra, we derive the following update rule: $\beta_k^j = \hat{\beta}_{k,-}^{j} + \hat{\beta}_{k,+}^{j}$ where
{\footnotesize
\begin{align}
\label{equ:update_rule}
\hat{\beta}_{k,-}^{j} &= \min\left[0,   \left( 1 +  
\sum_{j\in \bg} \frac{\lambda_2}{\left\|\bm{\beta}_k^{\bg}\right\|_2} +  
\sum_{k\in \bh} \frac{\lambda_3}{\left\|\bm{\beta}_{\bh}^{j}\right\|_2} \right)^{-1} \left\{\mathbf{r}_k (\mathbf{x}_j)^T+\lambda_1 \right\}\right],
\\ 
\hat{\beta}_{k,+}^{j} &= \max\left[0, \left( 1+  
\sum_{j\in \bg} \frac{\lambda_2}{\left\|\bm{\beta}_k^{\bg}\right\|_2} +  
\sum_{k\in \bh} \frac{\lambda_3}{\left\|\bm{\beta}_{\bh}^{j}\right\|_2} \right)^{-1} \left\{ \mathbf{r}_k (\mathbf{x}_j)^T-\lambda_1 \right\}\right].
\nonumber
\end{align}
}

We close this section by summarizing the desirable properties of our optimization method.
First, when $\bB$ is sparse, our optimization procedure is very fast.
We take advantage of not only the hierarchical structure of the DAG, 
but also the simple forms of the optimality conditions with residuals.
If we keep track of the residuals, we can efficiently check 
the optimality conditions for each sparsity pattern.
Second, our thresholding operations check all possible sparsity patterns,
resulting in appropriate structured sparsity in $\bB$.
It is important for eQTL mapping since
the coefficients for irrelevant SNPs can be set to exactly zero.
Third, our optimization method can deal with overlaps between/within 
the coefficients for input groups ($\bm{\beta}_k^{\bg}$'s)
and output groups ($\bm{\beta}_{\bh}^{j}$'s). 
Since input or output groups may overlap, and they must be considered simultaneously, 
this property of our method is essential. 
Finally, unlike some previous methods
\cite{yuan2006model,tibshirani1996regression},
we make no use of the assumption that the design matrix $\bX$ is orthonormal ($\bX^T\bX = \mathbf{I}$).
This dropping of the assumption is desirable for eQTL mapping in particular as
covariates (SNPs) are highly correlated due to linkage disequilibrium.
If one uses orthonormalization as a preprocessing step to make $\bX$ orthonormal,
there is no guarantee that the same solution for the original problem is attained \cite{friedman2010note}.

\section{Dealing with structures inducing higher-order effects}
\label{sec:model_interaction}

So far, we have been dealing with input and output structures in the context of multi-variate and multi-task linear regression where the influences from the covariates on the responses are additive. When higher interactions take place among covariates, which is known as epistasis and is prevalent 
in genetic associations \cite{carlson2004mapping}, a common approach to model such 
effects is polynomial regression \cite{montgomery2001introduction}, 
where higher-order terms of the covariates are included as additional regressors. However, in high-dimensional problems 
such as the one studied in this paper, this strategy is infeasible even for 2nd-order polynomial regression because, 
given say, even a standard genome dataset with $\sim 10^5$ SNPs, one is left with $\sim 10^{10}$ regressors 
which is both computationally and statistically unmanageable. In this section, we briefly show how to circumvent 
this difficulty using structured regularization based on prior information of covariate interactions. 
This strategy is essentially a straightforward generalization 
of the ideas in Section \ref{sec:methods} to a 
polynomial regression setting using a special type of structure encoded by a graph. 
Therefore all the algorithmic solutions developed in Section \ref{sec:optimization} 
for the general optimization problem in Section \ref{sec:methods} still apply here.  

Following common practice in GWA literature, here we consider only 2nd-order interactions between SNP pairs. 
Instead of including all SNP pairs as regressors, 
we employ a synthetic genetic interaction network \cite{costanzo2010genetic} 
to define a relatively small candidate set ${\bf U}$ of interacting SNP pairs. 
A synthetic genetic interaction network is derived from biological evidences of pairwise functional interactions between genes, 
such as double knockout experiments~\cite{tong2004global,koh2009drygin,costanzo2010genetic,boone2007exploring}. 
It contains information about the pairs of genes whose mutations affect the phenotype only when the mutations are present on both genes, 
and this represents a set of {\it ground-truth} interaction effects. 
Given such a network, we consider only those pairs of SNPs that are physically 
located in the genome near the genes that interact in the network within a certain distance. 
A 2nd-order regressor set ${\bf U}$ generated by this scheme is not only much smaller than an exhaustive pair-set, 
but also biologically more plausible.
Note that
it is possible to include other sets of SNP pairs from other resources in our candidate set. 
For example, in our experiments, we also added SNP pairs that passed two-locus epistasis test
with p-value $<10^{-5}$ into the set ${\bf U}$.

After finding the candidate SNP pairs, 
we generate the group of SNPs or interacting SNP pairs in two steps. 
In the first step, we find highly interconnected subgraphs (or clusters) 
from the genetic interaction network
using any graph clustering algorithms.
In our experiments, we used MCODE algorithm \cite{bader2003automated}  
for clustering the network.  
In the second step, we group all the SNPs or SNP pairs that are linked to the genes
in a cluster. We linked the genes and SNPs based on physical locations in the genome.
For example, if a SNP is located nearby a gene within a certain 
distance (e.g. $<$500bp), they are linked together.
Finally, we define individual SNPs in the $m$th group as ${\bg}_m \in \mathcal{G}$ 
and SNP pairs in the $m$th group as ${\bl}_m \in \mathcal{L}$.
We then look for associations between inputs/input-pairs and outputs via Eq. (\ref{equ:reg6}): 

\begin{subequations}
\label{equ:reg6}
\begin{align}
\min \frac{1}{2}
\sum_{k=1}^K\sum_{i=1}^{N} 
& \left( y_k^i - \sum_{j=1}^J{\beta_k^{j} x_j^i } -
\sum_{(r, s) \in {\bf U}}{\beta_k^{rs}  x_r^i x_s^i } 
\right)^2 
\label{equ:reg6_a}
\\
&  + \lambda_1   \sum_{k=1}^K \sum_{j=1}^J{|\beta_k^j|} 
\label{equ:reg6_b}     \\
  &  + \lambda_2  
\sum_{k=1}^K  \left(\sum_{m=1}^{|\mathcal{G}|}{\sqrt{\sum_{j\in {\bg}_m}({\beta_k^j})^2}} +
\sum_{m=1}^{|\mathcal{L}|}{\sqrt{\sum_{(r, s) \in {\bl}_m}{(\beta_k^{rs}})^2}}\right)
\label{equ:reg6_c}   \\
  & + \lambda_3  \left(\sum_{j=1}^J\sum_{m=1}^{|\mathcal{H}|} 
   \sqrt{\sum_{k \in {\bh}_m}{({\beta_k^j})^2}} +
   \sum_{(r, s) \in {\bf U}} \sum_{m=1}^{|\mathcal{H}|} 
   \sqrt{\sum_{k \in {\bh}_m}{({\beta_k^{rs}})^2}} \right)
  \label{equ:reg6_d}\\
&+ \lambda_4 \sum_{k=1}^K \sum_{\begin{subarray}{l}
       (r, s) \in {\bf U} 
       \end{subarray}}{|\beta_k^{rs}|}.
\label{equ:reg6_e}
\end{align}
\end{subequations}
where $\mathcal{G}$ is the set of input groups for 
marginal terms and $\mathcal{L}$ is the set of 
input groups for pairwise interaction terms.
Here, we use 
two tuning parameters for $L_1$ penalty
depending on whether a covariate is modeling an individual effect ($\lambda_1$) 
or interaction effect ($\lambda_4$) because
they might need different levels of sparsity.
Note that this problem is identical to Eq. (\ref{equ:reg5}) if we
treat interaction terms $x_r^i x_s^i$ as additional covariates, and hence
our optimization method presented in Section \ref{sec:optimization} is applicable to Eq. (\ref{equ:reg6}).  
However, Eq. (\ref{equ:reg6}) will be more computationally expensive than
Eq. (\ref{equ:reg5}) since Eq. (\ref{equ:reg6}) has
a larger number of covariates in $\bB$ including both marginal and interaction terms 
and additional tuning parameter $\lambda_4$.

\section{Simulation Study}
\label{subsec:simulstudy}

In this section we validate our proposed method using simulated genome/phenome datasets, and examine the effects of simultaneous use of input and output structures 
on the detection of true non-zero regression coefficients.
We also evaluate the speed and the performance of our optimization method
for support recovery in comparison to two other alternative methods.
For the comparison of optimization methods,
we selected smoothing proximal gradient method \cite{chen2010efficient} 
and the union of supports \cite{jacob2009group} since both methods
are in principle able to use input/output structures and handle overlapping groups.

\begin{figure}[htp]
\vspace{-0.5cm}
\centering
\hspace{-1.7cm}
\includegraphics[width=0.9\textwidth]{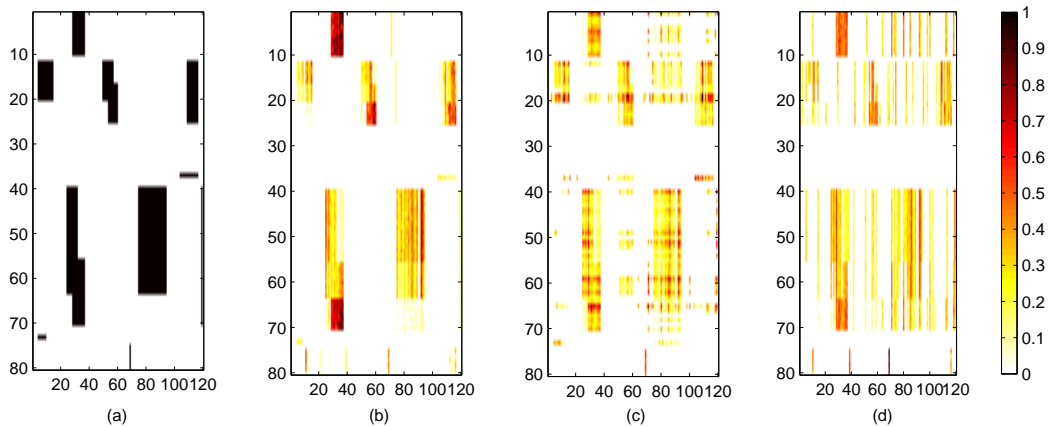}
\vspace{-0.5cm}
\caption{An example of simulation results with 
$|\beta_k^j| = 2, N=100, J=120$, and $K=80$. 
(a) True regression coefficient matrix.
Estimated $\bB$ 
by SIOL (b) with  both input and output structures
(c) with only input structure, and 
(d) with only output structure.
In 
(b-d), we show the normalized values of $|\bB|$.}
\label{fig:visualize_beta}
\vspace{-0.5cm}
\end{figure}

The simulated datasets with $J=120, K=80$, and $N=100$ are generated as follows.
For generating $\bX$, we first selected 60 input covariates from a uniform distribution over $\{0,1\}$
which indicates major or minor genotype. 
We then simulated 60 pairwise interaction terms $(x_j^i \times x_{j'}^i)$ by randomly selecting input-pairs from the 60 covariates mentioned above. Pooling the 60 marginal terms and 60 pairwise interaction terms resulted in a input space of 120 dimensions. 
We also defined input and output groups as follows (for the sake of illustration and comprehension  convenience, here our input and output groups correspond to variables to be jointly selected rather than jointly shrunk, the shrinkage penalty in our regression loss can be defined on the complements of these groups):

{\tiny
\begin{align}
& \rlap{$\overbrace{\phantom{5,\ldots,9,10}}^{\bg_1}$}
5,\ldots, \underbrace{9,10,\ldots,15}_{\bg_2}, \ldots,
\rlap{$\overbrace{\phantom{25,\ldots,29,30,31,32}}^{\bg_3}$}
25,\ldots,\underbrace{29,30,31,32,\ldots, 37}_{\bg_4}, \ldots,  
\rlap{$\overbrace{\phantom{50,\ldots,54,55,56,57}}^{\bg_5}$}
50,\ldots,\underbrace{54,55,56,57,\ldots,60}_{\bg_6}, \ldots,   
\rlap{$\overbrace{\phantom{75,\ldots,80,\ldots,87}}^{\bg_7}$}
75,\ldots,\underbrace{80,\ldots,87,\ldots,94}_{\bg_8}, \ldots,
\rlap{$\overbrace{\phantom{104,\ldots,109,110,111}}^{\bg_9}$}
104,\ldots,\underbrace{109,110,111,\ldots,116}_{\bg_{10}} \nonumber \\
& \rlap{$\overbrace{\phantom{1,\ldots,4,5}}^{\bh_1}$}
1,\ldots,\underbrace{4,5,\ldots,10}_{\bh_2}, \ldots,
\rlap{$\overbrace{\phantom{12,\ldots,17,18,19,20}}^{\bh_3}$}
12,\ldots,\underbrace{17,18,19,20,\ldots, 25}_{\bh_4}, \ldots,  
\rlap{$\overbrace{\phantom{46,\ldots,56,\ldots,63}}^{\bh_5}$}
46,\ldots,\underbrace{56,\ldots,63,\ldots, 70}_{\bh_6}, \ldots,  
\underbrace{75,\ldots,80}_{\bh_7}, \nonumber
\end{align}
}
where the numbers within a bracket represent the indices of
inputs or outputs for an input group $\bg_t,\; t=1,\ldots,10$, 
or an output group $\bh_o,\; o=1,\ldots,7$.
The inputs and outputs which did not belong to 
any groups were in a group by itself.
We then simulated $\bB$, i.e, the ground truth 
that we want to discover.  
We selected non-zero coefficients so that $\bB$ includes 
various cases, e.g., overlap between input and output groups, 
overlap within input groups, and overlap within output groups.
Figure \ref{fig:visualize_beta}(a) shows the simulated $\bB \in \mathbb{R}^{80 \times 120}$ where
non-zero coefficients are represented by black blocks.
Given $\bX$ and 
$\bB$, we generated $K=80$ outputs 
by $\bY=\bB \bX + \bE,$ 
$\bE \sim \mathcal{N}(\bm{0},\mathbf{I})$.
We generated 20 datasets and 
optimized Eq. (\ref{equ:reg5}) 
using the three methods.
We report the average performance using precision recall curves.

\subsection{Evaluation of the Effects of Using Input and Output Structures}
\label{subsec:simul_exp1}

We first investigate the effects of using both input and output structures on the performance of our model.
Here we applied our optimization method (HiGT) to the following three
models with different use of structural information: 
{\footnotesize
\begin{enumerate}
	\item Use of both input and output structures (Eq. (\ref{equ:reg5_b}) + Eq. (\ref{equ:reg5_c}) + Eq. (\ref{equ:reg5_d}))
	\item Use of input structures (Eq. (\ref{equ:reg5_b}) + Eq. (\ref{equ:reg5_c}))
	\item Use of output structures (Eq. (\ref{equ:reg5_b}) + Eq. (\ref{equ:reg5_d}))
\end{enumerate}
}

We then observed how the use of input/output structures affect the recovery of the 
true non-zero coefficients and the prediction error.
In Figure \ref{fig:visualize_beta}, we visualize the examples of 
estimated $\bB$ by the three different models.
Figure \ref{fig:visualize_beta}(b) shows that the model
with input and output structure successfully recovered true
regression coefficients in Figure \ref{fig:visualize_beta}(a).
However, as shown in Figure \ref{fig:visualize_beta}(c-d),
the models with either input or output structure 
were less effective to suppress noisy signals, which resulted in
many false positives.

\begin{figure}[htp]
\hspace{-1.7cm}
\includegraphics[width=1.2\textwidth]{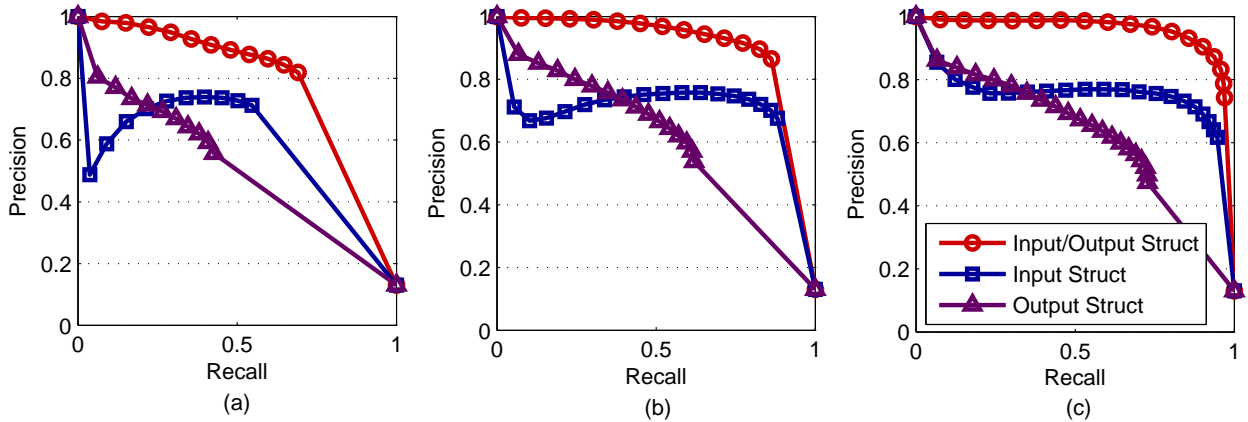}
\vspace{-1.5cm}
\caption{Precision recall curves on the recovery of true non-zero coefficients due to SIOL with both input and output structures (input/output struct), regression with only input structure (input struct), and with only output structure (output struct), under three different signal strengths of true regression coefficients.
(a) $\beta_k^j =0.4$, (b) $\beta_k^j =1$, and (c) $\beta_k^j =2$.
The simulated data were generated with $N=100, J=120$, and $K=80$. 
}
\label{fig:comp_case1}
\vspace{-0.5cm}
\end{figure}

Figure \ref{fig:comp_case1} shows the precision recall curves 
on the recovery of true non-zero coefficients by changing the threshold $\tau$ for choosing relevant covariates ($|\beta_k^j| > \tau$), 
under different signal strengths of 0.4, 1 and 2. 
For all signal strengths, the model with input/output structures 
significantly outperformed the other models with either input or output structure.
The most interesting result is that when the signal strength was very small such
as 0.4, our model still achieved good performance by taking advantage of both structural information.

\begin{figure}[htp]
\hspace{-0.3cm}
\centering
\subfigure[]{\includegraphics[width=0.3\textwidth]{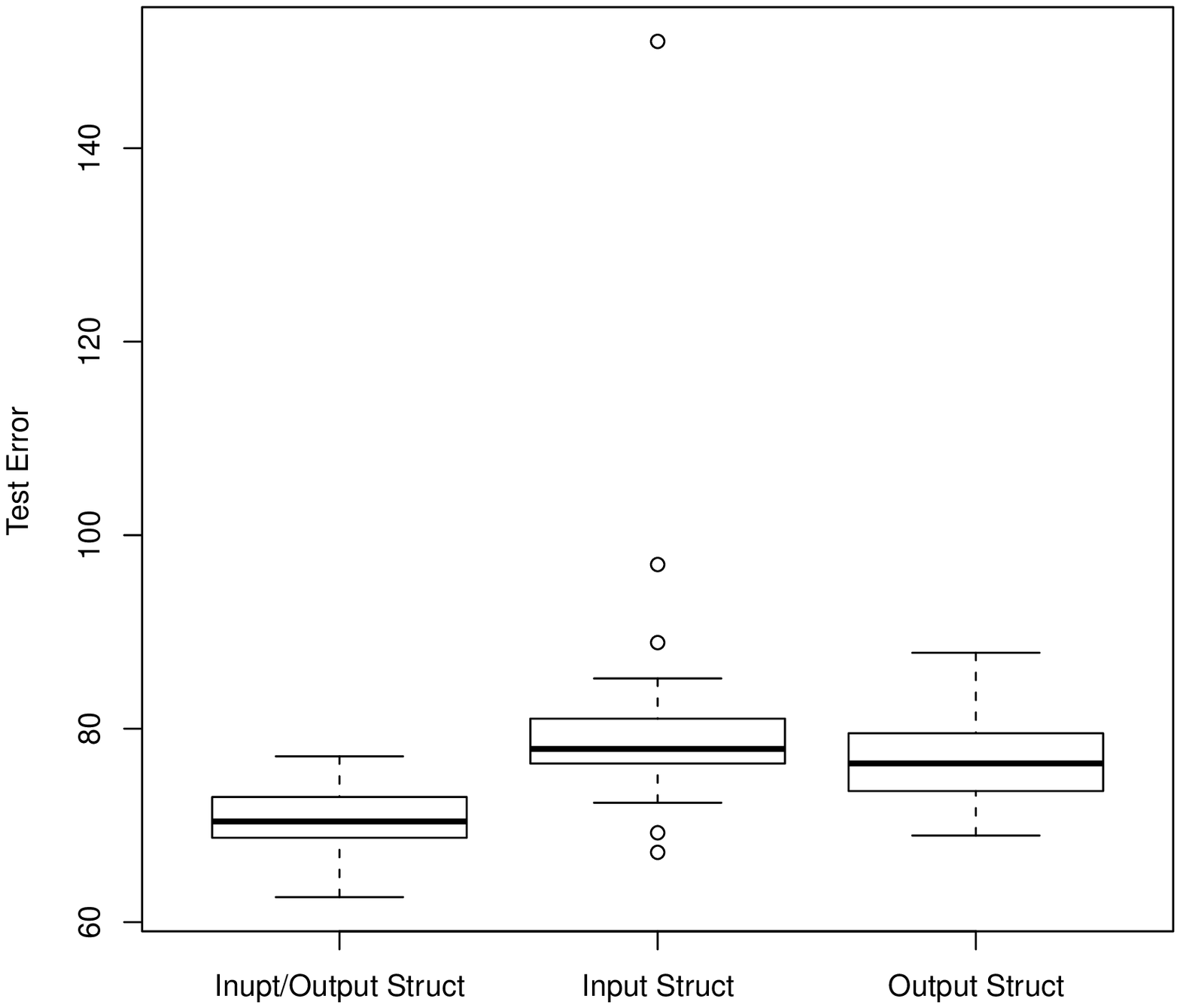}}
\subfigure[]{\includegraphics[width=0.3\textwidth]{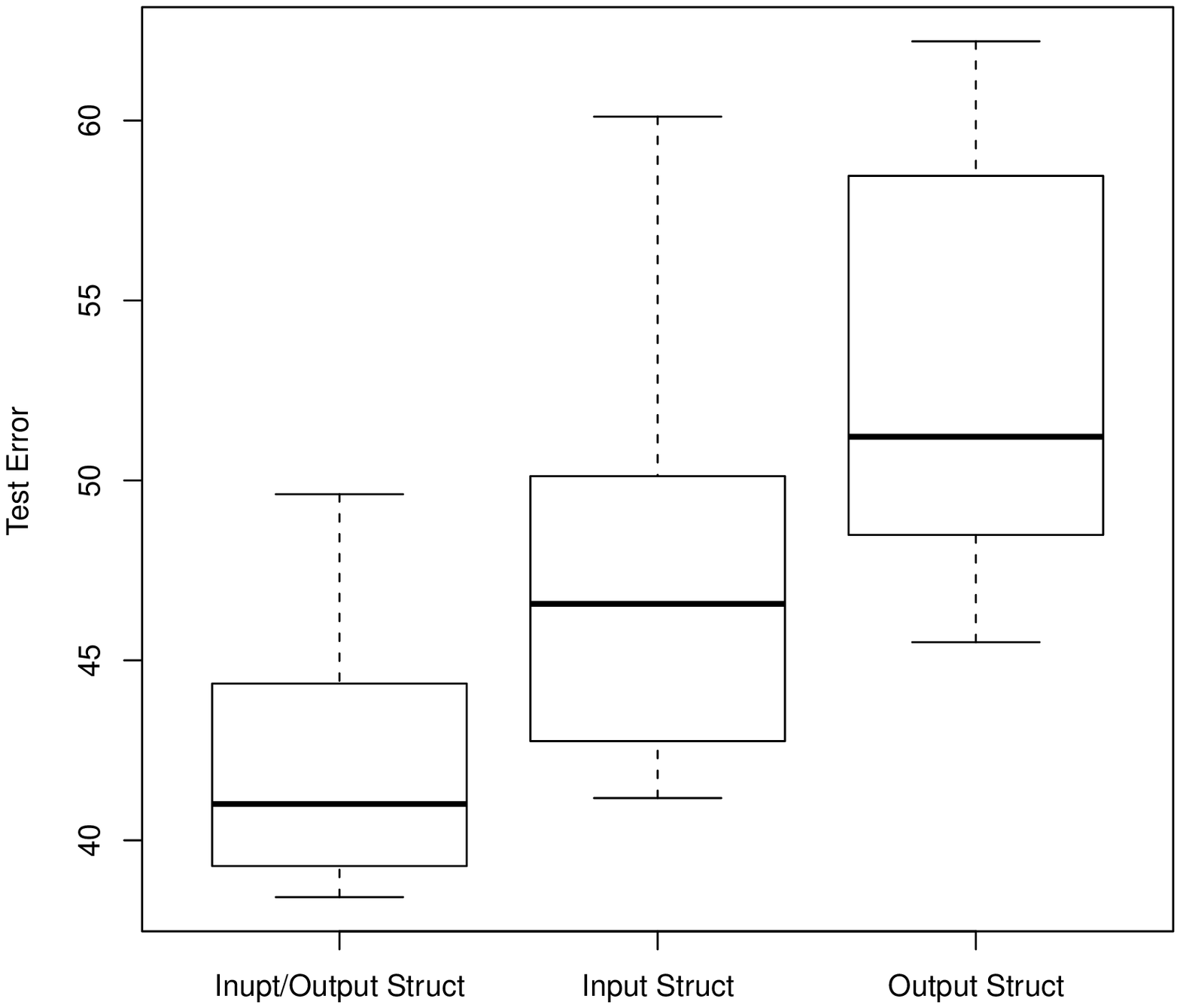}}
\subfigure[]{\includegraphics[width=0.3\textwidth]{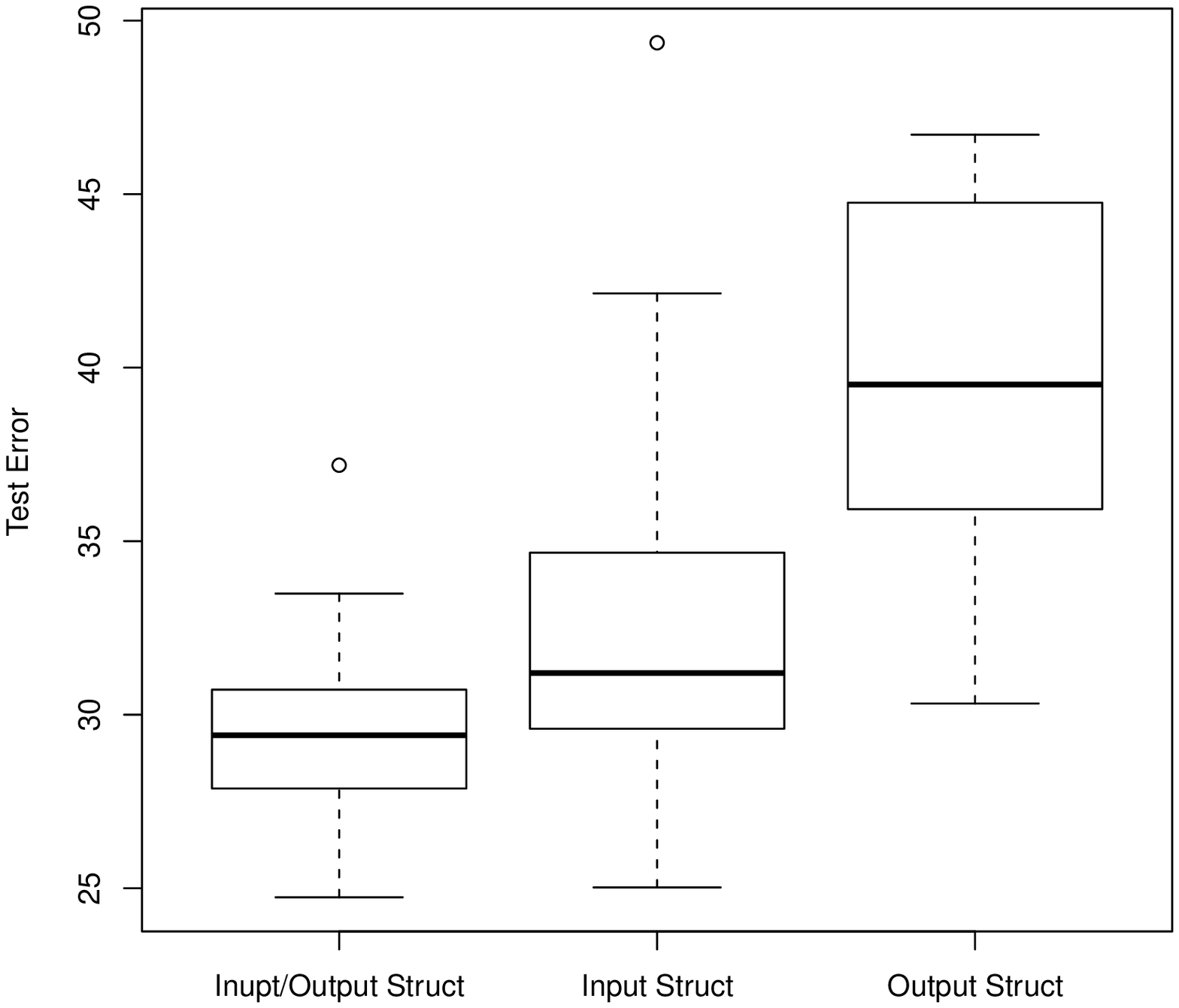}}
\vspace{-0.5cm}
\caption{Comparison of the prediction error of SIOL (input/output struct),
with regression under only input structure (input struct), on only output structure (output struct).
(a) $\beta_k^j = 0.4$, (b) $\beta_k^j = 1$, (c) $\beta_k^j = 2$.}
\label{fig:exp1_pred}
\vspace{-0.1cm}
\end{figure}

We also compare the prediction errors on our validation data with 280 ($20 \times 14$) samples 
(each dataset had 14 samples for validation).
For computing the prediction error, we first selected 
non-zero coefficients,
and then recomputed the coefficients of those selected covariates using linear regression without shrinkage penalty. 
Using the unbiased coefficients of the chosen covariates, we measured the prediction error for our 
validation data.
Figure \ref{fig:exp1_pred} shows the prediction error under different signal strengths ranging from 0.4 to 2.
For all signal strengths, we obtained significantly 
better prediction error using both input and output structures.
When the signal strength was large such as 1 or 2, the use of both input and output 
structures was especially beneficial for reducing the prediction error
since it helped the model to find most of the true covariates relevant to the outputs.

\begin{figure}[t]
\centering
\includegraphics[width=0.7\textwidth]{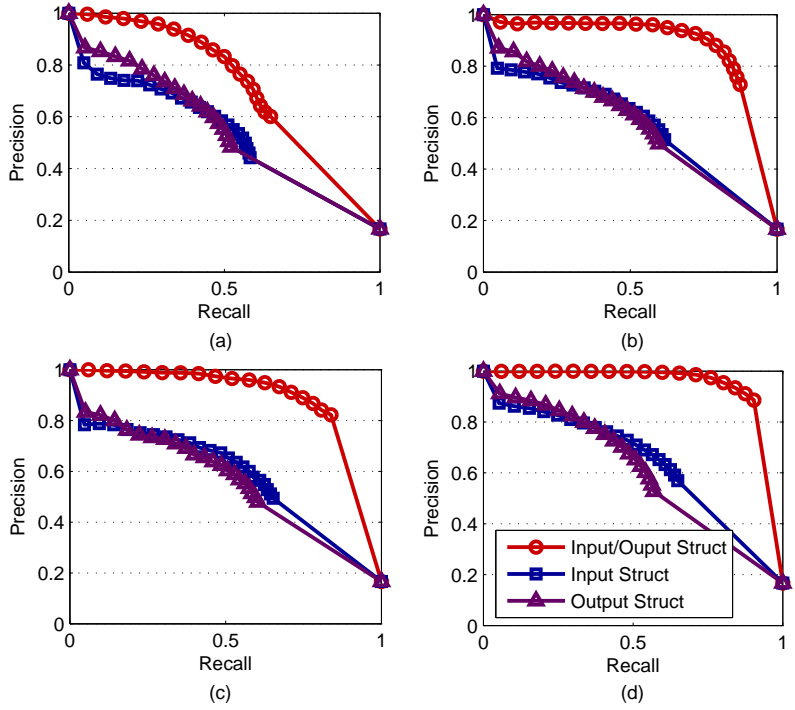}
\vspace{-0.5cm}
\caption{Precision recall curves on the recovery of true non-zero coefficients 
for SIOL (input/output struct), regression with only input structure, and with only output structure, 
under two different sizes of input and output groups.
(a) $|g| \in \{2,3\}$, 
(b) $|g| = 5, \; \forall g \in \mathcal{G}$, and
(c) $|h| = 5$, and 
(d) $|h| = 40, \; \forall h \in \mathcal{H}$.
We fixed the size of output groups for (a,b) ($|h| = 10$), and fixed
the size of input groups for (c,d) ($|g| = 5$).
The simulated data were generated with $\beta_k^j =0.5$, $N=100$, and $J=120$.
}
\label{fig:comp_case23}
\vspace{-0.3cm}
\end{figure}

\paragraph{The effects of the size of input and output groups}
Figure (\ref{fig:comp_case23}a-\ref{fig:comp_case23}d) demonstrates the results on
simulated datasets with different size of input and output groups.
For all group sizes, our method 
significantly improved the performance 
by effectively taking advantage of both input and output groups.

\subsection{Comparison of HiGT to Alternative Optimization Methods}
In this section, we compare the accuracy and speed of our optimization method (HiGT)  
with those of the two alternative methods including smoothing proximal gradient method (SPG) 
\cite{chen2010efficient}
and union of supports \cite{jacob2009group}.
Both alternatives can handle overlapping groups.
Specifically, the smoothing proximal gradient method is developed to efficiently deal with 
overlapping group lasso penalty and graph-guided fusion penalty using an approximation approach.
However, it may be inappropriate for our model since
the maximum gap between the approximated penalty and the exact penalty
is proportional to the total number of groups $R$, where 
$R = J  |\mathcal{H}| + K  |\mathcal{G}|$. 
Thus, when dealing with high dimensional data (e.g $J \sim 500,000$) such as genome data, 
the gap will be large, and the approximation method can be severely affected.
On the other hand, ``union of supports'' finds the support of $\bB$ from the union
support of overlapping groups.
To obtain the union of supports, input variables are duplicated to 
convert the penalty with overlap into the one with disjoint groups, and 
a standard optimization technique for group lasso \cite{yuan2006model} can be applied.
One of disadvantages of union of supports is that the number of duplicated input variables
increases dramatically when we have a large number of overlapping groups.
In our experiment, we considered all possible combinations of overlapping input and output groups,
and used a coordinate descent algorithm for sparse group lasso \cite{friedman2010note}.

\begin{figure}[htp]
\hspace{-1.7cm}
\includegraphics[width=1.2\textwidth]{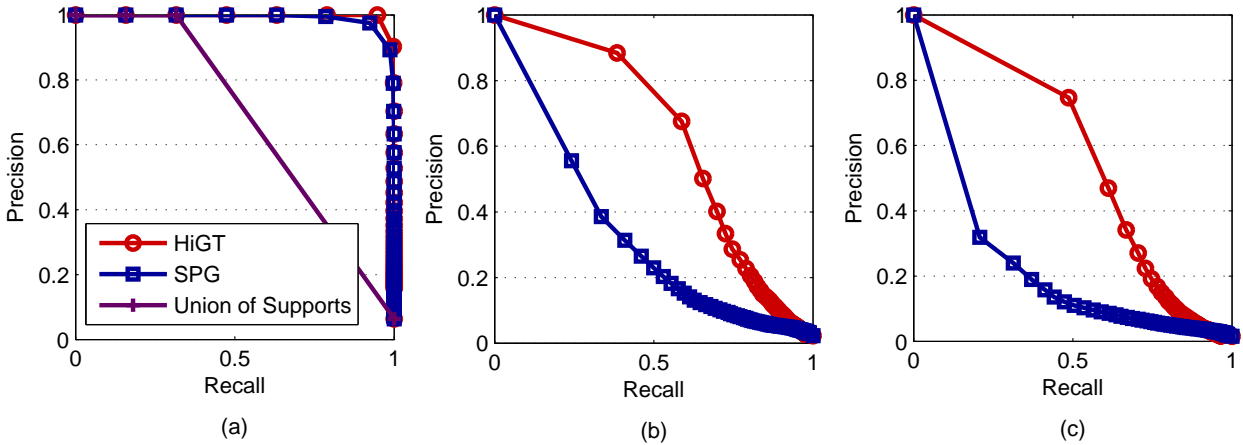}
\vspace{-1.5cm}
\caption{Precision recall curves 
on the recovery of true non-zero coefficients using the SIOL model via HiGT, smoothing proximal gradient method (SPG), and union of supports for optimization. Three different model sizes determined by the number of input variables were tested (due to high computational cost, results of union-of-support are only available for the smallest problem sizes tested): 
(a) $J = 30$, (b) $J = 400$, and (c) $J = 600$.
The simulated data were generated with $\beta_k^j =2$, $N=100$, and $K=20$.
}
\label{fig:accuracy}
\end{figure}

Figure \ref{fig:accuracy} shows the precision recall curves 
on the recovery of true non-zero coefficients under the SIOL model using the three optimization methods. The size of the problem is controlled by increasing number of input variables (from 30 to 600).
The simulated data set used here was identical to the data 
in Section \ref{subsec:simul_exp1}
except that we used 20 outputs ($\by_{61},\ldots, \by_{80}$) and different number of input variables.
One can see that our method outperforms the other alternatives for all
configurations. 
Our method and smoothing proximal gradient method 
showed similar performance when the input variable is small ($J=30$) 
but as $J$ increases, our method significantly performed better than SPG.
It is consistent with our claim for the maximum 
gap between the approximated penalty and the exact penalty which is
related to the number of groups.
Union of supports did not work well even when the number of 
input variables is small ($J=30$) since the actual number of 
input variables considered was very large due to the duplicated covariates, which
severely degraded the performance. 
\begin{figure}[htp!]
\vspace{-0.3cm}
\centering
\subfigure[]{\includegraphics[width=0.4\textwidth]{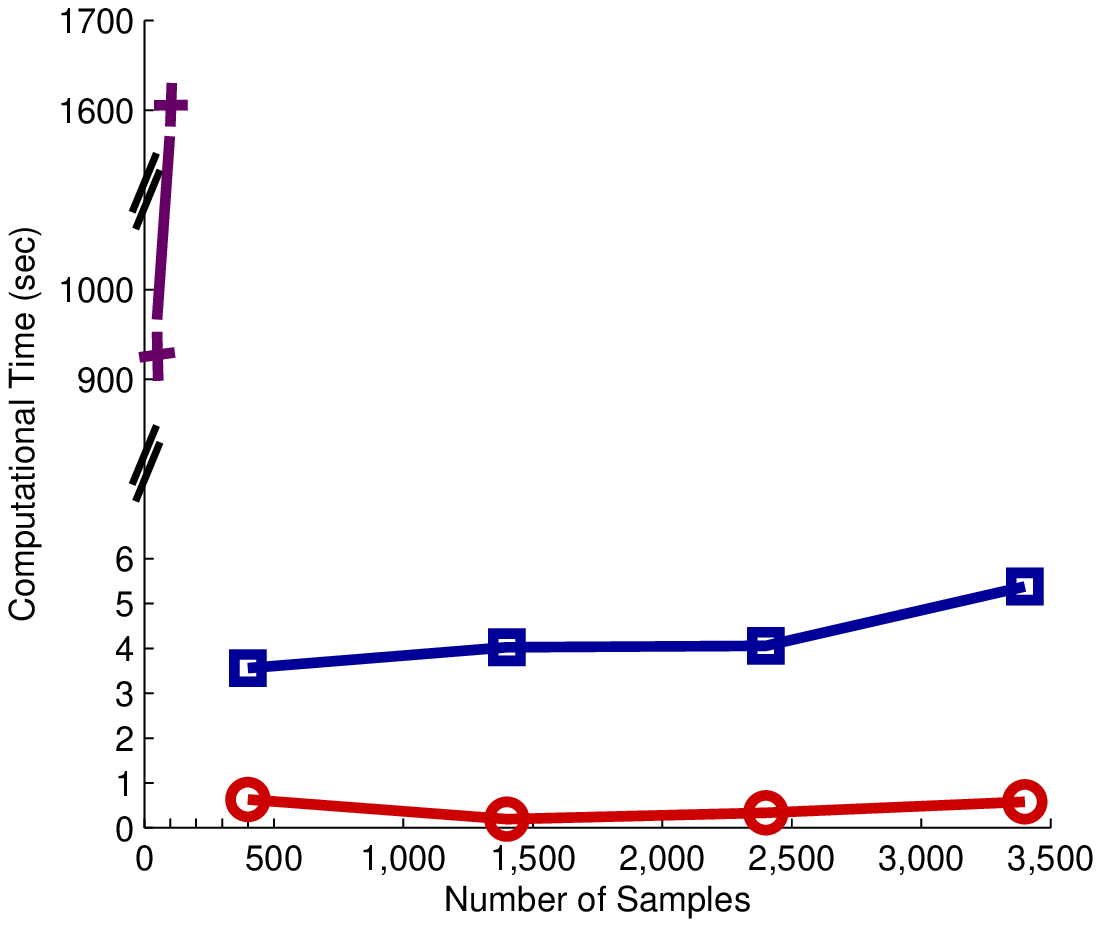}}
\subfigure[]{\includegraphics[width=0.4\textwidth]{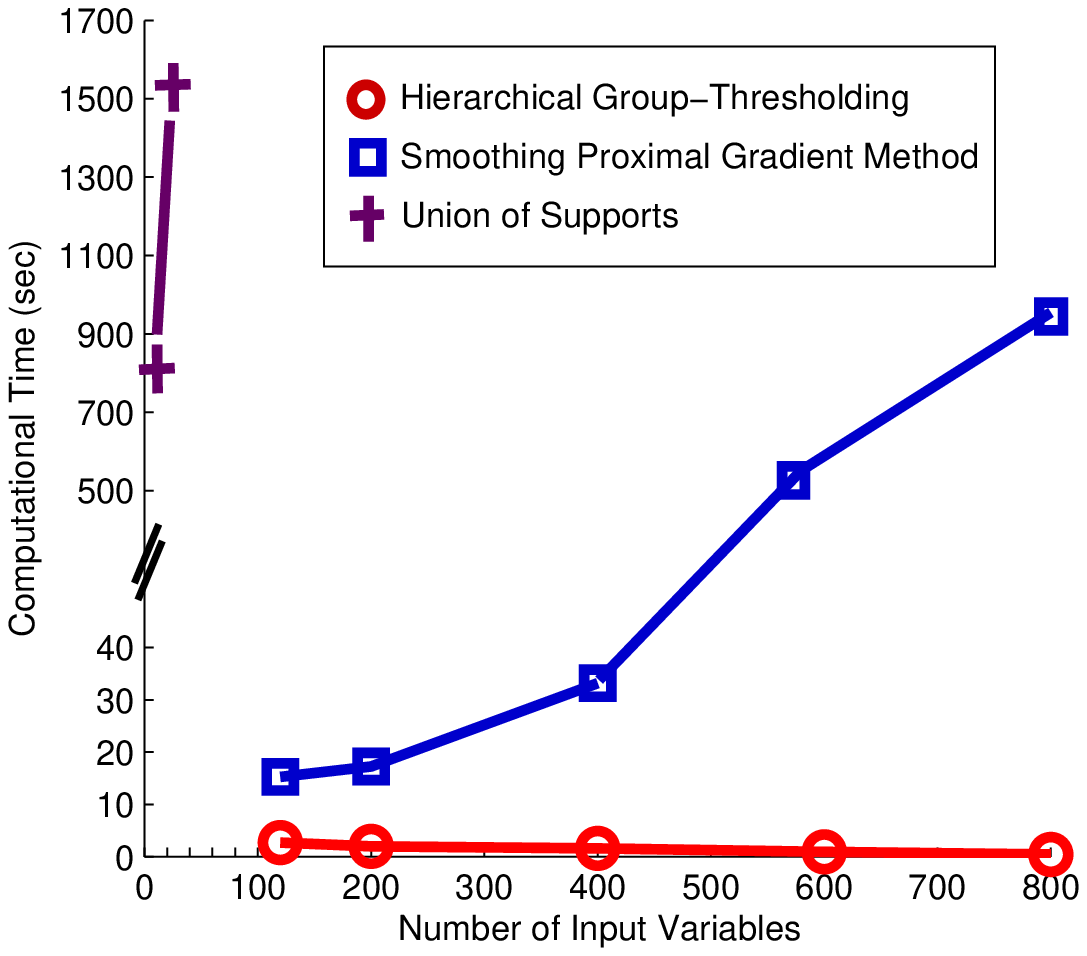}}
\vspace{-0.5cm}
\caption{Time complexity of HiGT, SPG, and union of supports.  
All three methods used both input and output groups. 
(a) Computational time with different number of samples, 
(b) computational time with different number of inputs. 
We used the same tuning parameters for all the methods ($\lambda_1 = 0.01, \lambda_2 = \lambda_3 = 0.1$).
We did not report the times for the small number of samples and inputs
for our method and SPG since I/O latency was dominant.
}
\label{fig:speed}
\vspace{-0.3cm}
\end{figure}

We also compared the speed of our method with the two alternatives
of union of supports with all possible combinations of input and output groups
and SPG that considered both input and output groups.
Figure \ref{fig:speed}(a,b) show that our method converged faster than
the other competitors, 
and was significantly more scalable than the two alternatives.
Union of supports was very slow compared to our method and SPG because
of the large number of duplicated input variables.
Our experimental results confirms that our optimization technique is not only accurate
but also fast, which
can be explained by the use of DAG and the simple forms of optimality checks.

\section{Analysis of Yeast eQTL Dataset}
\label{subsec:yeast_eqtl}
We apply our method
to the budding yeast (Saccharomyces cerevisiae) data 
\cite{brem2005landscape} with 1,260 unique SNPs (out of 2,956 SNPs)
and the observed gene-expression levels of 5,637 genes.
As network prior knowledge,
we used genetic interaction network reported in
\cite{costanzo2010genetic} with stringent cutoff to
construct the set of candidates of SNP pairs ${\bf U}$.
We follow the procedure in section \ref{sec:model_interaction}
to make ${\bf U}$ with an additional set of significant SNP pairs
with p-value $< 10^{-5}$ computed from two-locus epistasis test.
When determining the set ${\bf U}$,
we assumed that a SNP is linked to 
a gene if the distance between them is less than 500bp.
We consider it a reasonable choice for cis-effect as
the size of intergene regions for S. cerevisiae is 515bp 
on average \cite{sunnerhagen2006comparative}.
As a result, we included 982 interaction terms from 
the interaction network 
in $\bX$ 
with 1,260 individual SNPs.
The number of SNP pairs from two-locus epistasis test
varied depending on the trait.
For generating input structures, we processed the network data as follows.
We started with genetic interaction data 
which include 74,984 interactions between gene pairs. 
We then extracted genetic interactions with low p-values ($<$0.001).
Given 44,056 significant interactions, using MCODE clustering algorithm, we found 55
gene clusters. 
Using the gene clusters, we generated the groups of 
individual SNPs and pairs of SNPs according to the scheme in 
section \ref{sec:model_interaction}.
For generating output structures, we 
applied hierarchical clustering to the yeast gene expression data
with cutoff 0.8, resulting in 2,233 trait clusters.

\paragraph{Marginal Effects in Yeast eQTL dataset}

\begin{figure}[htp]
\centering
\includegraphics[width=0.9\textwidth]{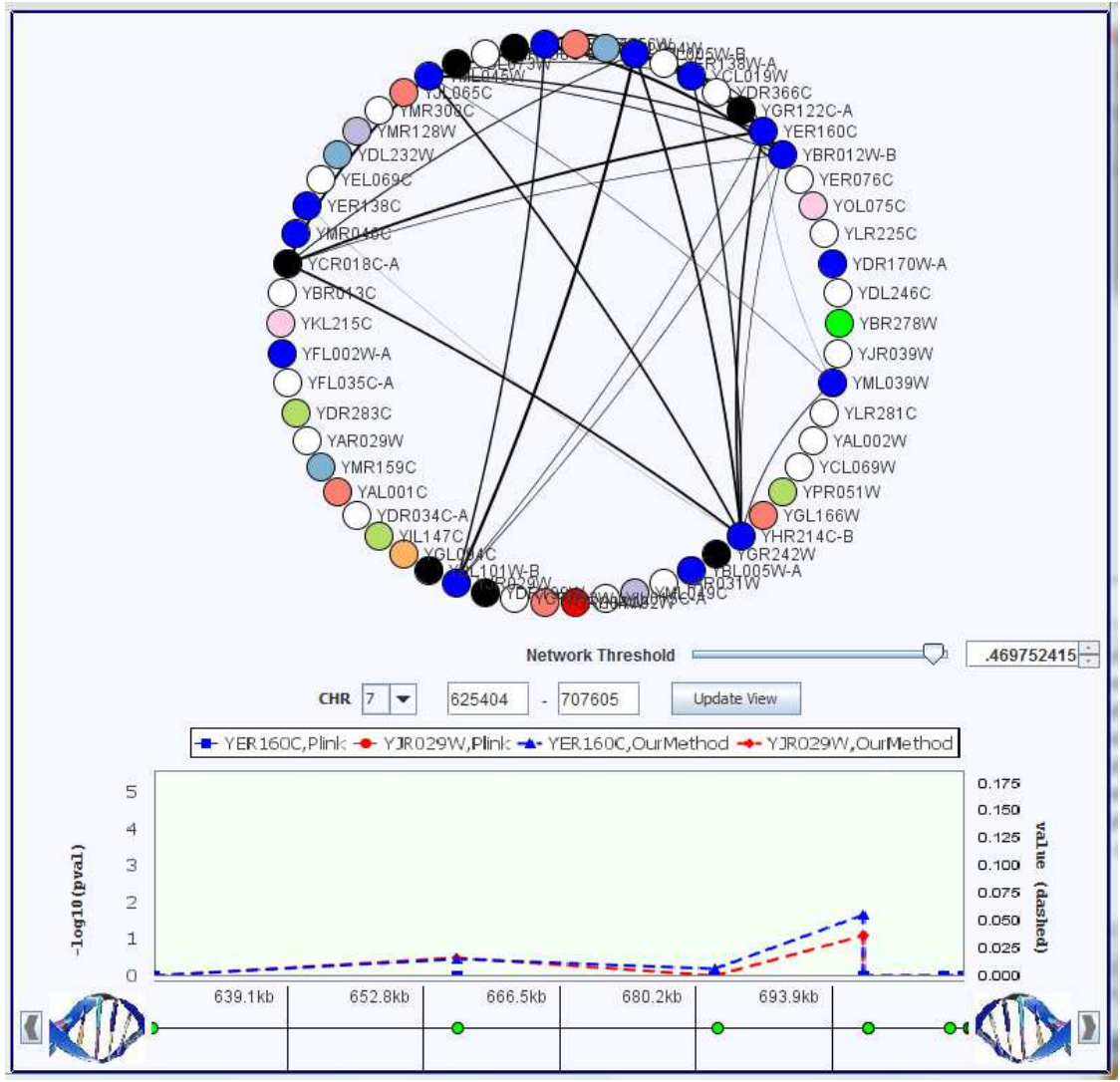}
\caption{Manhattan plot for association between (YER160C and YJR029W) and SNPs on chromosome 7.
The two genes YER160C and YJR029W share the same GO category ``transposition''.
Our method detected SNPs which affect both two genes in this region.
However, single SNP analysis did not find any associated SNPs and
lasso found SNPs  associated only with YER160C in this region. Graph were generated using the GenAMap software \cite{curtisgenamap}. 
}
\label{fig:marginal_diff}
\end{figure}
We briefly demonstrate the effects of input/output 
structures on the detection of eQTLs with marginal effects.
In general, the association results for marginal effects by 
our method, lasso and single SNP analysis 
(the later two are standard methods in contemporary GWA mapping 
that use no structural information,
and hence included for comparison) showed similar patterns for strong associations.
However, we observed differences
for SNPs with small or medium sized signals. 
For example, our results had fewer nonzero regression coefficients
compared to lasso. One possible explanation would be  
that the grouping effects induced by our model with input/output structures
might have removed
false predictions with small or medium sized effects.
To illustrate eQTLs with marginal effects, 
we show some examples of association SNPs using GenAMap \cite{curtisgenamap}.
Figure \ref{fig:marginal_diff} demonstrates a Manhattan plot on chromosome 7
for two genes including YER160C and YJR029W.
Both genes have the same GO category 
``transposition''.
As both genes share the same GO category,
it is likely that they are affected by
the same SNPs if there exist any association SNPs for both genes.
In our results, we could see that the same SNPs 
on chromosome 7 are associated with both genes
as shown in Figure \ref{fig:marginal_diff}. 
However, single SNP analysis did not find any significant
association SNPs in the region.
Lasso detected association SNPs in the region but 
they were associated with only YER160C rather than both of them
(lasso plot is not shown to avoid cluttered figure).
This observation is interesting since it supports that our method 
can effectively detect the SNPs 
jointly associated with the gene traits by taking advantage of
structural information.

\paragraph{Epistatic Effects in Yeast eQTL dataset}

\label{subsubsec:epi_yeast_eqtl}
\begin{figure}[htp]
\centering
\subfigure[]{\includegraphics[width=0.42\textwidth]{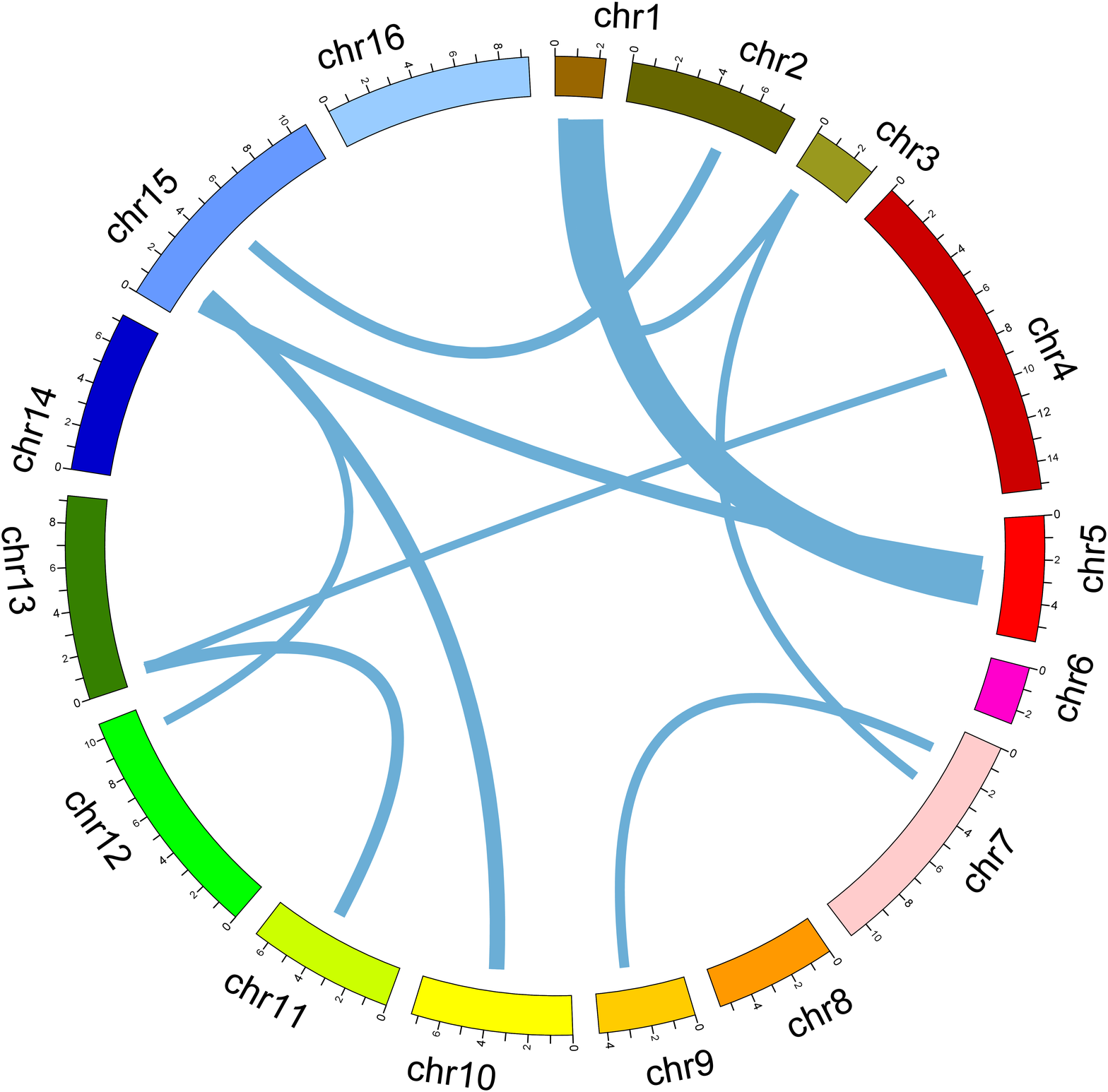}}
\subfigure[]{\includegraphics[width=0.42\textwidth]{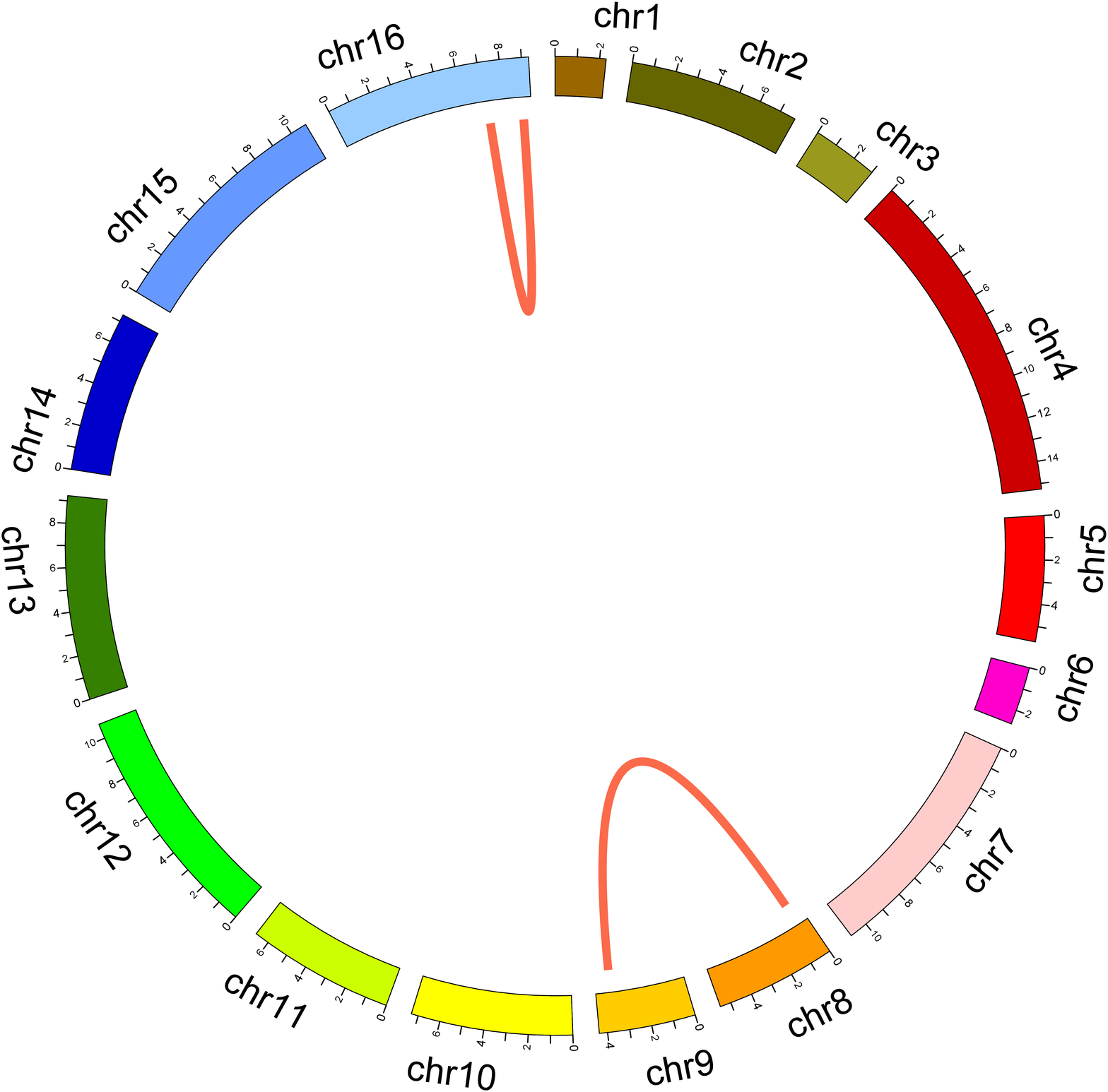}}
\caption{Hotspots with interaction effects identified by
(a) our method and (b) two-locus epistasis test. 
This figure represents the yeast genome
in a circular format. In clockwise direction, from the top
of the circles, we show 16 chromosomes, 
which are separated with space and different colors.
Lines indicate interaction effects
between two connected locations in the genome.
Thickness of the lines is proportional to the number of traits
affected by the interaction effects.
Here we show interaction effects which influence 
more than 100 gene traits. 
The hotspots for (a) are represented in Table \ref{tab:epistatic_hotspot}.
In (b), two SNP pairs are found including
chr16:718892-chr16:890898
(affected genes are enriched with the GO category of
ribosome biogenesis with corrected p-value $1.6\times10^{-36})$,
and chr8:56246-chr9:362631 
(affected genes are enriched with the
GO category of vacuolar protein catabolic process with corrected p-value $1.6\times10^{-14})$.
This figure was generated using Circos software \cite{krzywinski2009circos}.
}
\label{fig:epi_hotspots}
\vspace{-0.1cm}
\end{figure}

\begin{sidewaystable}
\caption{Hotspots of SNP pairs having epistatic effects in yeast identified by our method. 
}
\centering
\begin{tabular}{|c|c|c|c|c|c|c|c|}
\hline
Hotspot & SNP1  & SNP2  &  Number of  & GO category of  & Corrected p-value of \\ 
label & location &location &  affected traits & affected traits & GO category\\
\hline \hline
1  & chr1:154328 & chr5:350744  &  455& ribosome biogenesis & $1.2\times10^{-36}$\\
2  &chr10:380085 &   chr15:170945  &  195& ribosome biogenesis & $1.6\times10^{-12}$\\
3 &chr10:380085 &   chr15:175594  &  185 & ribosome biogenesis & $4.1\times10^{-12}$\\
4  &chr5:222998 &   chr15:108577  &   170& response to temperature stimulus & $2.9\times10^{-6}$  \\
5  &chr11:388373 &   chr13:64970  &   155& regulation of translation & $1.8\times10^{-32}$\\
6  &chr2:499012 &   chr15:519764  &   145& vacuolar protein catabolic process & $1.4\times10^{-7}$ \\
7  &chr1:41483 &   chr3:64311  &   130&  & \\
8  &chr7:141949 &  chr9:277908  &   125&  &\\
9  &chr3:64311 &   chr7:312740  &  115& glycoprotein metabolic process & $1.5\times10^{-4}$\\
10  &chr12:957108 &   chr15:170945  &   110 & vacuolar protein catabolic process & $7.8\times10^{-16}$\\
11 &chr4:864542 &  chr13:64970  &   105& ribonucleoprotein complex biogenesis & $3.7\times10^{-6}$\\
\hline
\end{tabular}
\label{tab:epistatic_hotspot}
\end{sidewaystable}

Now we show the benefits of using the input/output structures
for detecting interaction effects among SNPs by comparing the results of our method to those of two-locus epistasis test
performed by PLINK \cite{purcell2007plink} which uses no structural information.
Specifically, we compare the hotspots with interaction effects (i.e. SNP pairs that affect a 
large number of gene traits) which are identified by both methods.  
Recall that two-locus epistasis test is the most widely used 
statistical technique for
detecting interaction effects in genome-wide association studies, which
computes the significance of interactions 
by comparing between the null model
with only two marginal effects and the 
alternative model with two marginal effects and their 
interaction effect.
In the following analysis, we discarded all SNP pairs
if the correlation coefficient between the pairs $>0.5$ to avoid trivial 
interaction effects.

We first identified the most significant hotspots that 
affect more than 100 gene traits.
To make sure that we include only significant interactions,
we considered interaction terms 
if their absolute value of regression coefficients
are $> 0.05$. 
For the results of two-locus epistasis test,
we considered all SNP pairs with p-value $< 10^{-5}$.
Figure \ref{fig:epi_hotspots}(a,b) show the hotspots
found by our method and two-locus epistasis test. 
The rings in the figure represent the yeast genome
from chromosome 1 (located at the top of each circle) to 16 clockwise, 
and the lines show interactions 
between the two genomic locations at both ends.
One can see that our method detected 11 hotspots but
two-locus epistasis test found only two 
significant hotspots with interaction effects.
This observation shows that our method can find more 
significant hotspots with improved statistical power 
due to the use of the input/output structures.
In Table \ref{tab:epistatic_hotspot}, we summarized the 
hotspots found by our method.
It turns out that our findings are also biologically interesting
(e.g. 9 out of 11 hotspots showed GO enrichment).
Notably, hotspot 1 (epistatic interaction between 
chr1:154328 and chr5:350744) affects 455 genes which are enriched with 
the GO category of ribosome biogenesis with 
a significant corrected p-value $< 10^{-35}$
(multiple testing correction is 
performed by false discovery rate \cite{maere2005bingo}).
This SNP pair was included in our candidates from the genetic interaction network.
There is a significant genetic interaction between NUP60 and RAD51 
with p-value $3 \times 10^{-7}$ \cite{costanzo2010genetic}, and 
both genes are located at chr1:152257-153877 and
chr5:349975-351178, respectively.
As both SNPs are closely located to NUP60 and RAD51 (within 500bp), 
it is reasonable to hypothesize that 
two SNPs at chr1:154328 and chr5:350744 affected the two genes,
and their genetic interaction in turn acted on a large number of
genes related to ribosome biogenesis.

To provide additional biological insights,
we further investigated the mechanism of this significant SNP-SNP interaction.
From literature survey, RAD51 (RADiation sensitive) is a strand exchange
protein involved in DNA repair system \cite{sung1994catalysis}, 
and NUP60 (NUclear Pore) is the subunit of unclear pore complex involved in nuclear 
export system \cite{denning2001nucleoporin}. 
Also, it has been reported that yeast cells 
are excessively sensitive to DNA damaging agents if 
there exist mutations in NUP60 \cite{nagai2008functional}. 
In our results, we also found out that the SNP close to NUP60 
did not have significant marginal effects, and
the SNP in RAD51 had marginal effects.
According to these facts, it would be possible to hypothesize the following.
When there is no mutation in RAD51,
the point mutation in NUP60 cannot affect other traits since
the single mutation is not strong enough and 
if there exist DNA damaging agents in the environment,
DNA repair system would be able to handle them.
However, when there exist point mutations in RAD51, 
DNA damaging agents would  severely harm yeast cells with the point mutation in NUP60
since DNA repair system might not work properly due to the mutation in RAD51 
(recall that the SNP in RAD51 had marginal effects). 
As a result, 
both mutations in NUP60 and RAD51
could make a large impact on many gene traits.

\begin{figure}[htp]
\centering
\subfigure[]{\includegraphics[width=0.42\textwidth]{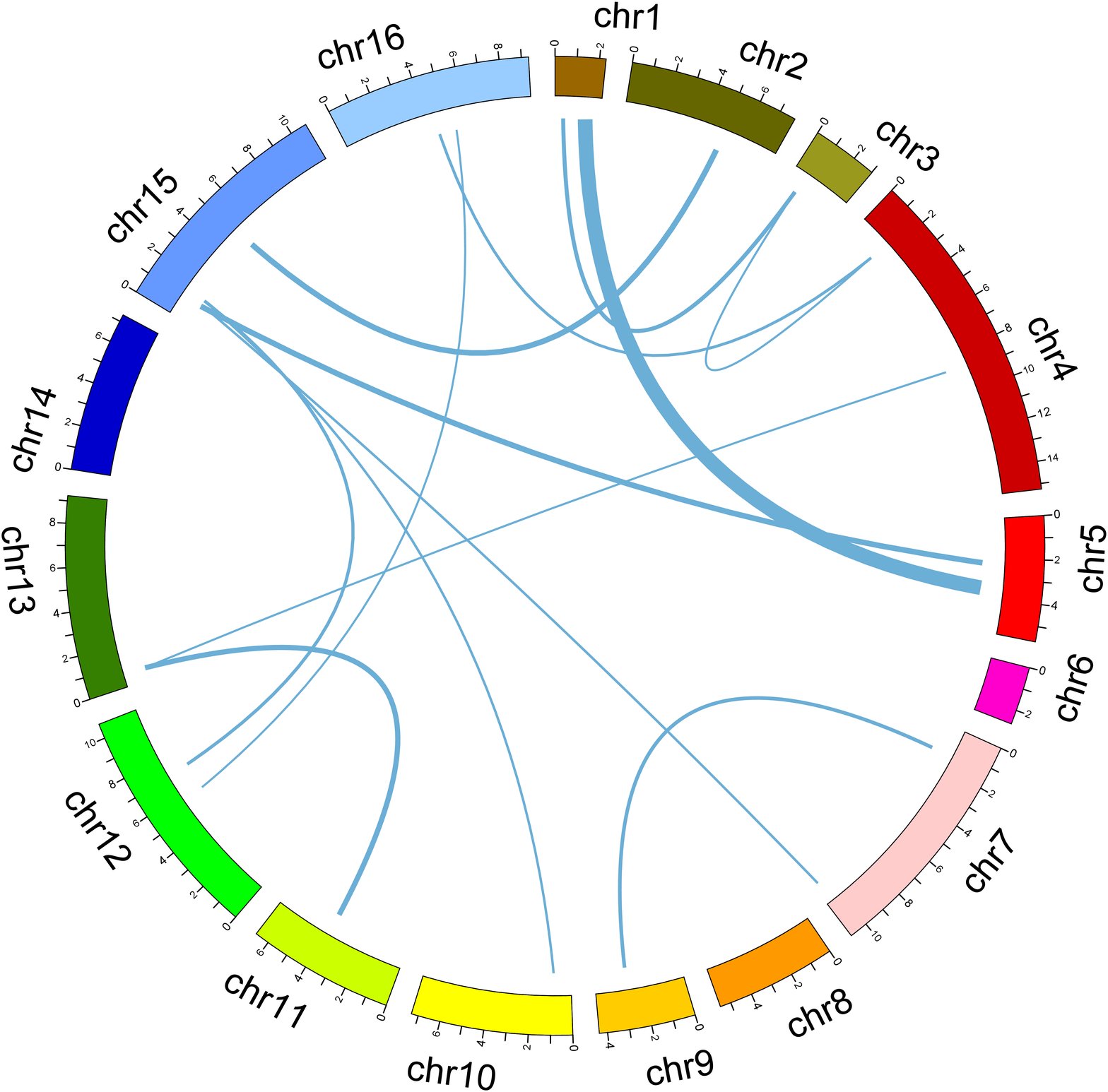}}
\subfigure[]{\includegraphics[width=0.42\textwidth]{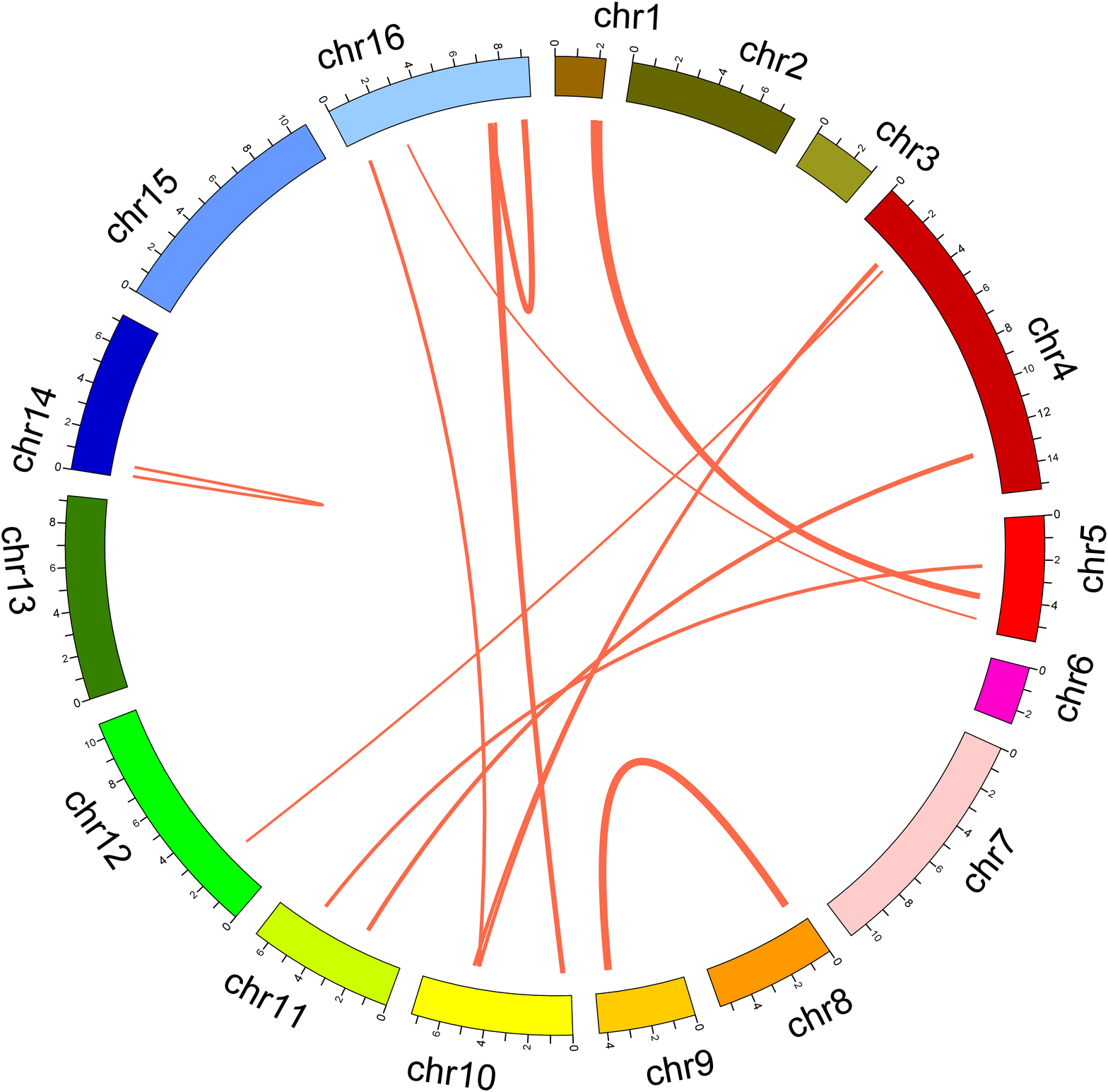}}
\caption{Hotspots with interaction effects identified by
(a) our method and (b) two-locus epistasis test by PLINK. 
Here we show epistatic interactions which influence 
more than 10 gene traits.
This figure was generated using Circos software \cite{krzywinski2009circos}.
}
\label{fig:medium_epi_hotspots}
\vspace{-0.1cm}
\end{figure}

\begin{figure}
\vspace{-0.5cm}
\centering
\includegraphics[width=0.45\textwidth]{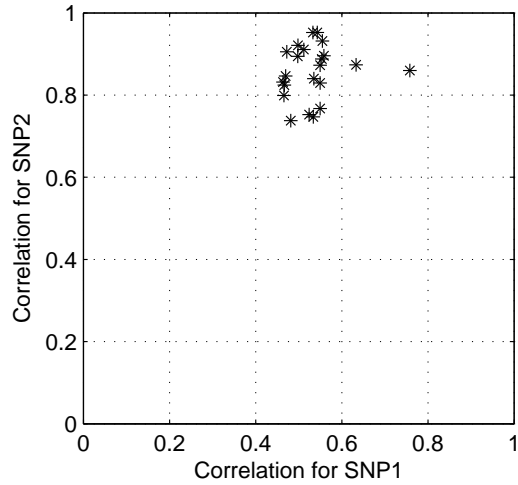}
\vspace{-0.3cm}
\caption{The scatter plot for illustrating the correlation 
between our epistatic hotspot 1 (chr1:154328-chr5:350744) and 
significant SNP pairs close to hotspot 1
detected by two-locus epistasis test
(p-value $< 10^{-6}$ and the distance between 
the pairs of SNPs and hotspot 1 is within $< 50$kb).
Each dot represents a SNP pair (SNP1, SNP2) 
found by two-locus epistasis test,
and x-axis represents the correlation between SNP1 and chr1:154328
and y-axis represents the correlation between 
SNP2 and chr5:350744. 
Each dot was perturbed by a small amount of random noise
to avoid overlapping of the dots. 
}
\label{fig:corr_graph}
\vspace{-0.5cm}
\end{figure}

\begin{figure}[htp]
\vspace{-0.5cm}
\centering
\subfigure[]{\includegraphics[width=0.4\textwidth]{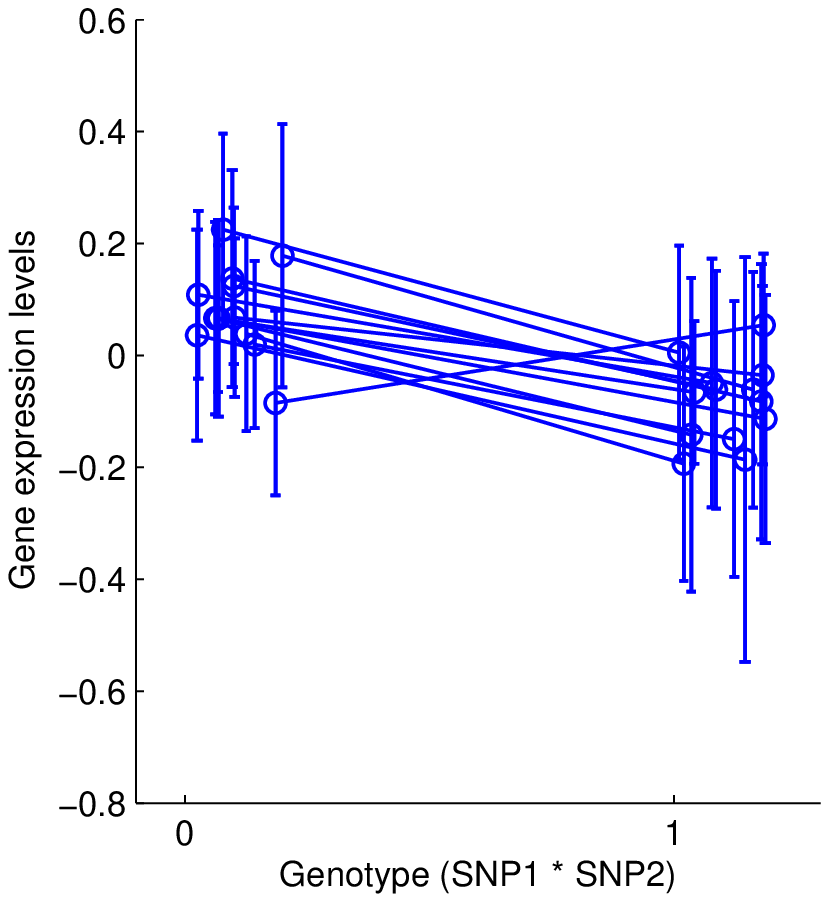}}
\subfigure[]{\includegraphics[width=0.4\textwidth]{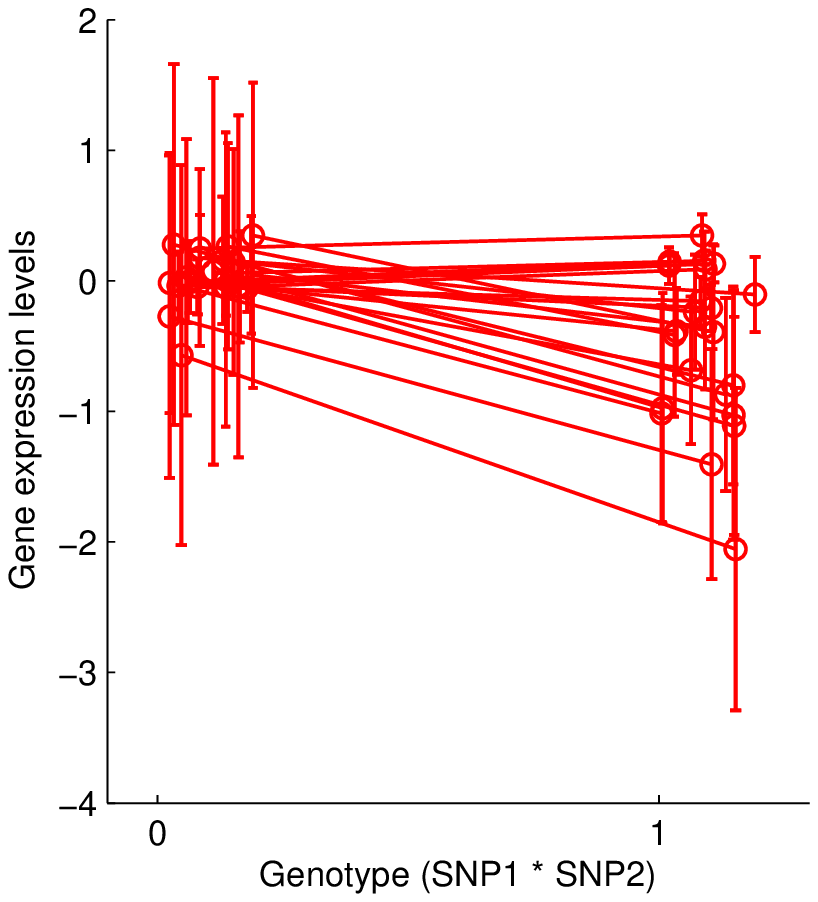}}
\vspace{-0.5cm}
\caption{Variations of gene expression levels w.r.t. to the 
genotypes of (a) a SNP pair (chr10:87113-chr15:141621)
found by our method, and 
(b) a different SNP pair (chr8:63314-chr9:362631) 
found by two-locus epistasis test.
Here, x-axis represents genotype doses and
y-axis shows the average expression levels of the multiple genes (denoted by multiple vertical lines) affected by the corresponding SNP pairs. 
A small noise was added to the genotypes as offsets to avoid overlapping of the error bars.}
\label{fig:checks}
\vspace{-0.8cm}
\end{figure}

Furthermore, we looked at the hotspots which affect $>10$ gene traits.
Figure \ref{fig:medium_epi_hotspots}(a,b) show epistatic interactions 
identified by our method and two-locus epistasis test, respectively.
In this figure, we show significant interactions with 
regression coefficient cutoff $>0.1$ for our method, and
p-value cutoff $<10^{-6}$ for two-locus epistasis test.
These cutoffs are arbitrarily chosen to make the number of hotspots
found by both methods similar.
Surprisingly, two methods showed very different hotspots with 
epistatic interactions. 
Figure \ref{fig:medium_epi_hotspots}(a) was very similar to 
Figure \ref{fig:epi_hotspots}(a) but
in Figure \ref{fig:medium_epi_hotspots}(b), 
several hotspots emerged which were absent in 
Figure \ref{fig:epi_hotspots}(b).
We will analyze these hotspots in two ways.
First we will look at the hotspots with epistatic effects 
which appeared in both Figure \ref{fig:medium_epi_hotspots}(a) and (b).
Then we will investigate the differences between the two
results.
First, we observed that both methods 
found significant epistatic effects between chromosome 1 and 5.
Recall that in our previous analysis of the hotspots, 
this interaction was discussed
(see hotspot 1 in Table \ref{tab:epistatic_hotspot}).
Among all significant SNP pairs found by two-locus epistasis test,
there was no identical SNP pair to hotspot 1
but there were 30 SNP pairs close to it (within $<50$kb). 
Also, it turns out that these 30 SNP pairs had very strong correlation 
with hotspot 1. In Figure \ref{fig:corr_graph}, we show scatter plot to illustrate
the strong correlations between hotspot 1 and these 30 SNP pairs.
More interestingly, the total number of affected traits by these 30 SNP pairs
was 416, and it is very similar to 455 that is the number of affected 
genes by hotspot 1.
According to these facts and our previous analysis
for the mechanism of hotspot 1, it seems that hotspot 1 is 
truly significant, and two-locus epistasis test found 
significant SNP pairs that are close to the true location but 
failed to locate the exact location of hotspot 1.
It supports that our algorithm could find such a significant 
hotspot affecting $>$ 400 genes by detecting exact SNP pairs.
However, two-locus epistasis test was unable to 
locate many hotspots affecting a large number of traits 
due to insufficient statistical power.
Second, we investigated the differences between the two 
results in Figure \ref{fig:medium_epi_hotspots}(a,b). 
As we cannot report all the results in the paper,
we focused on a SNP pair (chr10:87113-chr15:141621) 
in Figure \ref{fig:medium_epi_hotspots}(a),
and another SNP pair (chr8:63314-chr9:362631) 
in Figure \ref{fig:medium_epi_hotspots}(b).
Figure \ref{fig:checks}(a,b) 
show the average gene expression levels for each SNP pair.
In this figure, x-axis represents the genotype $\in \{0,1\}$ 
which is the multiplication of two SNPs 
(SNP1 $\times$ SNP2, where SNP1, SNP2 $\in \{0,1\}$), and y-axis represents the
average gene expression levels 
of individuals with given genotype.
Each line in Figure \ref{fig:checks}(a,b) shows
how the average gene expression level varies as the genotype changes 
from 0 to 1 for each trait affected by the SNP pairs 
with error bars of one standard deviation.
Interestingly, in Figure \ref{fig:checks}(a), 
we could 
see that there is 
a consistent pattern,
where for most gene traits,
the expression levels decreased as the genotype changed from 0 to 1.
However, as shown in Figure \ref{fig:checks}(b),
for the SNP pair found by two-locus epistasis test,
we could not find such a coherent pattern.
It demonstrates the differences between our method and two-locus
epistasis test. As our model borrows information across input and output group structures, 
we could find consistent gene expression patterns for the SNP pair.
On the other hand, two-locus pairwise test analyzed each SNP pair
separately, and each trait affected by the SNP pair showed 
different gene expression patterns with different  standard deviations.
Thus, 
it seems that our method can provide interesting biological insights in terms of
gene expression patterns in addition to the statistical significance.

\section{Discussions}

In this paper, we presented jointly structured input-output lasso to 
simultaneously take advantage of both input and output structures.
We also presented an efficient optimization technique for solving our 
multi-task regression model with structured sparsity.
Our experiments confirmed that our model is able to significantly improve 
the accuracy for detecting true non-zero coefficients using both input and output
structures. 
Furthermore, we demonstrated that our optimization method is faster and more accurate than
the other competitors.
In our analysis of yeast eQTL datasets, we identified 
important pairs of eQTL hotspots that potentially interact with each other.

\paragraph{Prior knowledge about input and output structures} 
In practice, it is important to generate reliable input and output groups to maximize
the benefits of the structural information.
In our experiments, we showed that yeast genetic interaction networks 
can be used as prior knowledge to define input and output structures.
However, such reliable prior knowledge 
cannot be easily attained when we deal with human eQTL datasets.
Instead, we have a variety of resources for human genomes
including protein-protein interaction networks and pathway database.
Generating reliable input and output structures exploiting multiple resources
would be essential for the successful discovery of human eQTLs.

\paragraph{Comparison between HiGT and other optimization algorithms}
Recently, an active set algorithm \cite{jenatton2009structured} has been proposed developed for variable selection with structured sparsity, which can potentially be used for estimating the SIOL model. 
We observe two key differences 
between our method and the active set algorithm \cite{jenatton2009structured}.
First, the active set algorithm incrementally 
increases active sets by searching available non-zero patterns, hence
one can see that it is a ``bottom-up'' approach.
On the other hand, our method adopts ``top-down'' approach where
irrelevant covariates are discarded as we walk 
through the DAG. 
Second, our algorithm guarantees  to search all zero patterns
while the active set algorithm needs a heuristic to select candidate
non-zero patterns. When $\bB$ is sparse, our
algorithm is still very fast by taking advantage of the structures of DAG. 
However, when $\bB$ is not sparse, 
our algorithm needs to search a large number of zero patterns and
update many non-zero coefficients but
the active set algorithm still does not need to 
update many non-zero coefficients.
Hence, in such a non-sparse case, the active set algorithm may have 
an advantage over our optimization method.
Other alternative optimization methods for SIOL include
MM (majorize/minimize) algorithm \cite{lange2004springer}
and generalized stage-wise lasso \cite{zhao2007stagewise}.
However, these methods did not work well for SIOL as 
the approximated penalty by MM algorithm 
and the greedy procedure of generalized stage-wise lasso 
were incapable of efficiently inducing complex sparsity patterns.

\paragraph{Future work}
One promising research direction would be to systematically estimate the
significance of the covariates that we found.
For example, computing p-values of our results would be helpful
to control the false discovery rate. 
To handle both sparse and non-sparse $\bB$, 
it would be also interesting to develop an optimization method for our model
that can take advantage of both ``bottom-up'' and ``top-down'' strategies. 
For example, we can select variables using ``bottom-up''
approach and discard irrelevant variables using
''top-down'' approach alternatively in a single framework.
Finally, we are interested in applying our method to human disease datasets.
In that case, the extension of our work to handle case-control studies and
finding reliable structural information will be necessary.

\bibliographystyle{plain}

\end{document}